\documentclass[preprint,floatfix,superscriptaddress]{revtex4-1}

\makeatletter
\renewcommand{\p@subsection}{}
\renewcommand{\p@subsubsection}{}
\makeatother

\usepackage[utf8]{inputenc}
\usepackage{comment}
\usepackage[colorlinks,linkcolor=blue,citecolor=blue,urlcolor=blue]{hyperref}
\usepackage{amsmath,amsfonts,amssymb}
\usepackage{cleveref}
\usepackage{graphicx}
\usepackage{subfigure}
\usepackage{xcolor}
\usepackage{multirow}
\usepackage[titletoc]{appendix}
\usepackage{titlesec}

\begin{document}

\title{Bayesian SegNet for Semantic Segmentation with Improved Interpretation of Microstructural Evolution During Irradiation of Materials}

\author{Marjolein Oostrom}
\affiliation{Pacific Northwest National Laboratory, Richland, WA, USA}
\author{Alex Hagen}
\affiliation{Pacific Northwest National Laboratory, Richland, WA, USA}
\author{Nicole LaHaye}
\affiliation{Pacific Northwest National Laboratory, Richland, WA, USA}
\author{Karl Pazdernik}
\thanks{Corresponding author: marjolein.oostrom@pnnl.gov}
\affiliation{Pacific Northwest National Laboratory, Richland, WA, USA}
\affiliation{North Carolina State University, Raleigh, NC, USA}
\maketitle

\section{Abstract}

Understanding the relationship between the evolution of microstructures of irradiated LiAlO\textsubscript{2} pellets and tritium diffusion, retention and release could improve predictions of tritium-producing burnable absorber rod performance. Given expert-labeled segmented images of irradiated and unirradiated pellets, we trained Deep Convolutional Neural Networks to segment images into defect, grain, and boundary classes. Qualitative microstructural information was calculated from these segmented images to facilitate the comparison of unirradiated and irradiated pellets. We tested modifications to improve the sensitivity of the model, including incorporating meta-data into the model and utilizing uncertainty quantification. The predicted segmentation was similar to the expert-labeled segmentation for most methods of microstructural qualification, including pixel proportion, defect area, and defect density. Overall, the high performance metrics for the best models for both irradiated and unirradiated images shows that utilizing neural network models is a viable alternative to expert-labeled images.

\textbf{Keywords}: Deep Convolutional Neural Networks; segmentation; microstructures; tritium; irradiation

\section{Introduction}

The National Nuclear Security Administration’s Tritium Sustainment Program irradiates tritium-producing burnable absorber rods (TPBARs) to produce tritium, a component of nuclear weapons \cite{senor2018science}. Tritium production is affected by tritium diffusion, retention and release within the irradiated LiAlO\textsubscript{2} pellets within the TPBARs, and the relationship between the evolution of microstructures within the pellets and tritium diffusion, retention, and release is a current area of research \cite{jiang2020quantitative} \cite{senor2018science} \cite{jiang2022microstructural}. These microstructures include defects such as voids, zirconia impurites and LiAl\textsubscript{5}O\textsubscript{8} precipitates, as well as grain boundaries. Quantifying microstructural evolution during irradiation is an important step towards predicting tritium-producing burnable absorber rod performance.

Previous studies have provided descriptions of changes in pellet defects after irradiation. These defects include voids, zirconia impurities, and LiAl\textsubscript{5}O\textsubscript{8} precipitates, which are formed when Li-6 is used to produce tritium in Li-deficient areas. For example, Jiang et al. noted large precipitates in irradiated pellets \cite{jiang2022microstructural}. An earlier study also noted that precipitates were partly surrounded by voids in some irradiation conditions and found a higher density and size of voids after irradiation of the LiAlO\textsubscript{2} pellets \cite{jiang2020quantitative}. As noted in Jiang et al., the interaction of defect formation during the irradiation process with tritium retention and release warrants further study \cite{jiang2022microstructural}. Another important characteristic of the pellets is the grain boundary location and proportion, as the complex tritium release process involves grain boundary diffusion \cite{jiang2020quantitative}. Previous studies found the defect-denuded zones are present after both ion and neutron irradiation, and that some grain boundaries had an increased number of voids \cite{jiang2024microstructural}. Characteristics of these defects and boundaries, including proportions and spatial location patterns, are important to measure as inputs for tritium-producing burnable absorber rod performance prediction.
 
Segmenting images into defects, boundaries, and backgrounds is the first step in quantifying microstructures and material features in LiAlO\textsubscript{2} pellets. As described in Pazdernik et al., Deep Convolutional Neural Networks (DCNN)'s have previously been successfully utilized to segment images of unirradiated pellets of LiAlO\textsubscript{2} pellets, and the segmentation results were used to calculate material proportion and microstructural features  \cite{pazdernik2020microstructural}. DCNN models have also been used in segmenting micrographs in other materials such as in steel \cite{roberts2019deep}, irradiated FeCrAl alloys \cite{jacobs2022performance}, and perovskite solar cells \cite{li2024machine}. Pixel-by-pixel expert hand-labeling of LiAlO\textsubscript{2} pellets is labor-intensive, so using DCNN segmentation increases the feasiblity of the microstructure analysis of the pellets. In this paper, updated code from the previous study by Pazdernik et al. was applied on both unirradiated and irradiated pellet images. Using the expert-labeled pellet images of irradiated and unirradiated pellets, we compared several different model architectures to find the best model for predicting microstructures. We compared the microstructural quantities of the unirradiated and irradiated pellet images and also compared model-predicted with the expert-labeled microstructures quantities.

\section{Material and methods}

\subsection{SEM Imaging}
The unirradiated pellets were first sectioned with a slow-speed wafer saw and mounting in epoxy resin. The pellets were polished with sequentially finer glycol-based polishing material, concluding with suspensions of colloidal SiO\textsubscript{2}.

The unirradiated samples were imaged with a JEOL 7001FFEG scanning electron microscope. For the images of unirradiated pellets, the imaging conditions were the following: 30 Pa Low Vacuum mode, 1-2 nA beam current, 5.0 KV accelerating voltage, and 4.6 mm working distance. The images have a 44.9 \(\mu m\) height and 59.8 \(\mu m\) width.

The irradiated samples were created from LiAlO\textsubscript{2} pellets which were irradiated for 18 months at the Watts Bar Nuclear reactor in Tennessee, USA. The pellets were removed from the rods in a hot cell facility, and then sectioned with diamond saw, mounted in epoxy resin, and polished to a 0.005 \(\mu m\) SiO\textsubscript{2} finish.

The irradiated samples were imaged with a FEI Quanta 250FEG scanning electron microscope with a backscattered electron detector and an EDAX Electron Back-scattered Diffraction system. The imaging process was modified to handle the irradiated samples - for example, silver paint was added to improve the electrical connection to ground in the scanning electron microscope. The recorded meta-data for imaging irradiated pellets are shown in Table \ref{tab:irradiated_meta-data}. Additional meta-data is in Appendix Table  \ref{tab:additional_meta_data}. There are differences in resolutions between images so all values are converted to \(\mu m\) before being compared. The recorded value range for beam current ranges from 0.000142 to 0.000178, and we assumed the units of the recorded values are \(\mu\)A's so the beam current remained within the expected pico to nano range. However, the recorded beam current may be inaccurate because instrument configurations can affect the actual beam current without updating the recorded beam current.

\begin{table}[!ht]
    
    \caption {\label{tab:irradiated_meta-data}  Information  for irradiated pellet images. } The imaging conditions for irradiated pellet images 3-5. 
    \begin{tabular}{|l|l|l|l|l|}
    \hline
        Values & Unit & ``Image 3'' & ``Image 4'' & ``Image 5'' \\ \hline
        Low Vacuum & Pa & 40 & 30 & 40 \\ \hline
        Beam Current & pA & 158 & 178 & 142 \\ \hline
        Accelerating Voltage & KV & 10 & 5 & 15 \\ \hline
        Working Distance & mm & 7.3 & 5 & 5 \\ \hline
        Height & \(\mu m\) & 24.7 & 24.7 & 17.2 \\ \hline
        Width & \(\mu m\) & 16.1 & 23.8 & 19.9 \\ \hline
    \end{tabular}
    
\end{table}

\subsection{Input Pre-Processing}

We followed the same general methods for input pre-processing as was described in greater detail in Pazdernik et al. \cite{ pazdernik2020microstructural}. First, images were cropped if they included the edge of the sample. (The width in Table \ref{tab:irradiated_meta-data} is the width of the images after cropping). Each image was then divided width-wise into two sections, with one section containing the left three-fourths of the image-width, with the section-width rounded down to a multiple of 128. The left three-fourth of the original image was used to create the training and validation datasets and the remaining section was used as the test image. The training and validation section was divided in 128 pixel-wide square chips and the chips were split into 80\% and 20\% proportions of training chips and validation chips, respectively. The training dataset was augmented by all permutations of flipping the chips on the horizontal, vertical, and/or diagonal axis, which augmented the training dataset by a factor of eight $(2^3=8)$. The validation dataset was not augmented and the test subsection was not divided or augmented. 

\begin{figure}[htbp]
    
    \includegraphics[width=\textwidth]{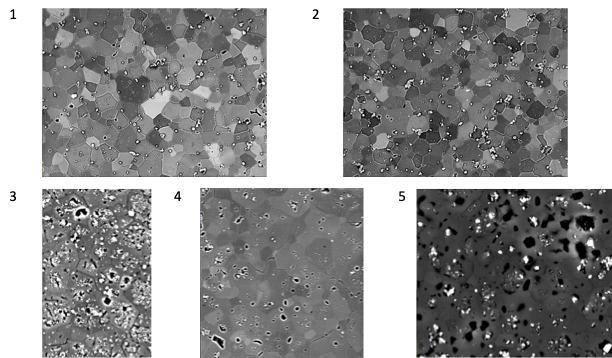}
    \caption{\label{fig:images_unlabeled} Unlabeled Images: All the unlabeled images used within the paper. Images 1-2: Unirradiated pellet images. Images 3-5: Irradiated pellet images. Images are scaled to the same height for illustrative purposes.}
    
\end{figure}

\begin{figure}[htbp]
    
    \includegraphics[width=\textwidth]{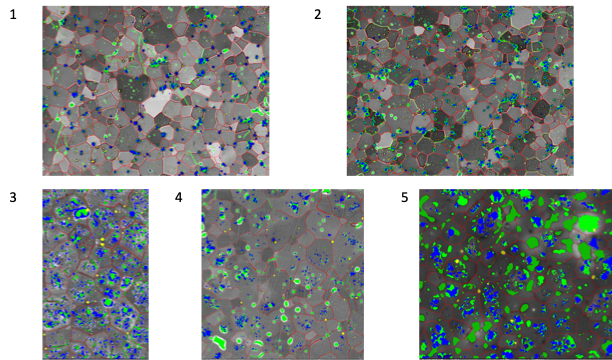}
    \caption{\label{fig:images_labeled} Labeled Images: All the labeled images used within the paper. Images 1-2: Unirradiated pellet images. Images 3-5: Irradiated pellet images. The color coding is as follows: grain (no labels), grain boundary (red), void (green), impurity (yellow), precipitate (blue). Images are scaled to the same height for illustrative purposes.}
    
\end{figure}

All of the expert-labeled data was converted from a colored label to an corresponding pixel class. The unlabeled and labeled images are in Figure \ref{fig:images_unlabeled} and in Figure \ref{fig:images_labeled}, respectively. The classes were grains, boundaries, impurities, precipitates, and voids, where the impurities, precipitates, and voids were all classified as defects. Three new irradiated pellet images and one new unirradiated pellet image were labeled in addition to the previously labeled image in Pazdernik et al. \cite{ pazdernik2020microstructural}. The labels of the unirradiated image created by Pazdernik et al. were updated to increase the thickness of the boundaries class to increase the accuracy of the expert-labels. 

\subsection{Model Architecture}

We decided to use DCNN architectures rather than Vision Transformers (ViTs) such as Segment Anything Model (SAM) \cite{kirillov2023segment} and Segment Everything Everywhere with Multi-modal (SEEM) \cite{zou2024segment} when training with our small dataset. Previous papers have noted that, while ViTs perform better than the DCNN architecture ResNet, when trained with large datasets (14M-300M images), ViTs performs worse than ResNet when trained on mid-sized datasets such as ImageNet (14M images) \cite {he2016deep} \cite{deng2009imagenet} \cite{dosovitskiy2020image}. Even with later modifications to the ViT architecture to improve performance on small training datasets, ViTs and ResNet perform comparably when trained from scratch in image classification tasks \cite{liu2021efficient}. Considering that we also have a small dataset, the DCNN architecture seemed to be appropriate for our training data size. 

We also considered using pretrained ViT models, but SAM performs poorly on medical images due to the lack of medical images in the SAM training data  \cite{huang2024segment}. (We were unable to find any information on the performance of SEEM with medical images \cite{zou2024segment}.) Our data shares many of characteristics that make medical images different from the SAM training data - our images are also gray-scale, with unclear boundaries, unclear foreground and background distinction, and with very small objects \cite{huang2024segment}. Early attempts to use these model on our images did not result in useful segmentation.

We used SegNet \cite{badrinarayanan2017segnet}, the DCNN architecture that resulted in a better performing model for unirradiated pellet images than the ResNet and  U-net models architectures in Pazdernik et al. \cite{badrinarayanan2017segnet}, \cite{ pazdernik2020microstructural} \cite{he2016deep} \cite{ronneberger2015u}. In the SegNet architecture, the decoder uses pooled indices from the corresponding encoder so the model does not need to learn to up-sample \cite{badrinarayanan2017segnet}. It is designed to maintain class boundaries with a relatively low computational and memory cost. The efficiency of SegNet was important considering the number of models we trained to compare a large number of parameter combinations, including model architectures, loss weights, and optimizers \cite{badrinarayanan2017segnet}. We used the same SegNet architecture as previously implemented in Pazdernik et al. \cite{ pazdernik2020microstructural}, using the code located at the ``pytorch-segnet'' GitHub repository \cite{say4n}, which is an implementation of the SegNet architecture from Badrinarayanan et al. \cite{badrinarayanan2017segnet}. The SegNet model was initiated with pretrained VGG-16 model weights in the encoding layers.

In addition to SegNet \cite{badrinarayanan2017segnet}, we tested modifications to the SegNet model. The first architecture modification, the ``Bayesian SegNet'' model architecture, adds a dropout layer during both model training and testing to a basic SegNet architecture \cite{kendall2015bayesian}. Drop-out layers, layers with neurons that are randomly dropped from the model's network, are commonly used to prevent over-fitting \cite{srivastava2014dropout}. Another advantage of using drop-out layers during inference is the changes in the layers between trials can result in different predictions, and the distribution of the predictions can be used to better measure uncertainty \cite{kendall2015bayesian}. Bayesian SegNet used a SegNet model implementation with a Bayesian dropout rate of 50\% in both the training and testing \cite{Ganaye}.

The second architecture modification, the ``Small Bayesian SegNet'' model architecture, is a modification of Bayesian SegNet with less encoding and decoding layers than the regular Bayesian SegNet. We used a smaller model to see if we could lower the computational cost with a smaller model without a decrease in performance. In addition, a smaller model is less likely to overfit. Small Bayesian SegNet uses only 10 encoder layers and 9 decoder layers rather than the default 13 encoder layers and 12 decoder layers. 

The last architecture modification, the ``Bayesian SegNeSt'' model architecture, incorporates a split attention block, as used in ResNeSt \cite{zhang2022resnest}, as well as the Bayesian dropout layers. ResNeSt has achieved a high accuracy for image classification tasks compared with the concurrent state-of-the-art models \cite{zhang2022resnest}. In a split attention block, the model splits into K feature map groups, where K is the cardinality hyperparameter, and each feature map group is further split into R groups, where R the radix hyperparameter \cite{zhang2022resnest}. We used a radix hyperparameter of 2. We trained the Bayesian SegNet and SegNeSt models without initiating the weights with a pretrained model. These models incorporated a 50\% drop-out rate into the SegNet architecture which makes the pre-trained weights less applicable for the modified architecture. For all the Bayesian models, we used the SegNet implementation from the ``pytorch-unet-segnet'' GitHub repository \cite{trypag} and used the default arguments from Hagen et al.\cite{hagen2022dbcal}.

With each of these model architectures, we tested two weights for the loss function. First, we tested weighted cross entropy (WCE), where the segmentation class is weighted by the inverse of its proportion. Secondly, we tested expert-weighed cross entropy (EWCE), where the grain class was weighed ten-fold less than the other classes. This was based on a subject-matter expert's opinion on the relative importance of each class. We also tested two optimizers, Adam and RMSprop, to optimize the gradient descent during the model training.
 
\subsection{Meta-data Informed Model}

After training models with all combinations of parameters - model architectures, WCE or EWCE weights for the loss function, and optimizers - we selected the best parameters separately for unirradiated and irradiated pellets by comparing the trained model predictions with the expert predictions. We also tested further altering the architectures and loss weights to improve the performance metrics of the models. 

First, we tested incorporating meta-data into the model architecture to decrease the model's sensitively to imaging conditions. In a study by Borges et al., imaging information was incorporated into a segmentation model to reduce the effect of the image acquisition method on segmentation performance \cite{borges2019physics}. In that study, a sequence of parameters were used to create simulated Magnetic Resonance Imaging images and the images were segmented with a neural network. The parameters were incorporated by concatenating the parameters at the final encoder to decoder shortcut of the 3D U-Net architecture. In our study, the  parameters included any image meta-data that was likely to affect the contrast or size of the image, including beam voltage, image field width, and image magnification. We concatenated the meta-data after the last encoding layer of the SegNet architecture so the meta-data would be incorporated in the sequential decoding network.

To create the meta-data informed model, meta-data values were normalized across images for numerical values and converted to dummy values for categorical values. The meta-data was only available for irradiated data and can be seen at Appendix Table  \ref{tab:additional_meta_data}. The meta-data was included as an input to the SegNet model architecture after the last encoding layer. We chose to incorporate the meta-data into the SegNet model as it was the best performing model in Pazdernik et al. \cite{pazdernik_microstructural_2020}. 

\subsection{Focal Loss}

It's essential for the model to detect defects, which have a large class imbalance with the grain background. To improve the detection of the defects, we implemented focal loss. The focal loss should strongly incentivize learning from incorrect predictions, and this should increase the model’s performance with difficult classes. In a focal loss, the parameter `gamma' multiplies one minus the raw score (0-1) of an correct label exponentially.

The focal loss function uses a `gamma' value to determine how much a well-classified point is down-weighted when calculating the loss. A `gamma' value of 0 leads to no down-weighting. A standard `gamma' value of two (2) results in a loss function where poorly classified predictions (which have a low score for the correct class) are weighted more heavily than well classified predictions (which have a high raw score for the correct class) \cite{lin2017focal}. We used a `gamma' value of 2, which was the optimal value used in the paper on focal loss  \cite{lin2017focal}. We used a multi-class implementation of focal loss as described in Lin et al.  \cite{lin2017focal} located at the ``pytorch-multi-class-focal-loss'' GitHub repository \cite{AdeelH}.

\subsection{Topological Loss}

Another loss function we tested was the topological loss function from Hu et al. \cite{hu2019topology}, which we utilized to improve the boundary predictions. In this loss function, the ground truth and a likelihood map created by the neural network are compared to each other using persistent homology. Persistent homology records the appearance (birth) and disappearance (death) of topological structures in a persistence diagram as the threshold for a likelihood map increases from 0 to 1. The loss function minimizes the distance between the persistence diagram of the likelihood map and of the ground truth \cite{hu2019topology}. In our paper, we have five possible classes rather than the binary class used as the example in Hu et al., so we used only the boundary class to create both our likelihood map and ground truth.

The topological loss value was added to the cross-entropy loss after 475 epochs, as the notes in the code repository associated with Hu et al. recommended adding the topological loss only after the model had been trained for a number of epochs \cite{hu2019topology}. The training and validation accuracy across epochs in Pazdernik et al. shows that the accuracy has stabilized by 475 epochs \cite{ pazdernik2020microstructural}. To convert our data into the binary data needed for the topological loss function, we first converted the model output with the softmax function across the class dimensions, and then we used the softmax values from the boundary class as our likelihood map and used the boundary class from the labeled images as the ground truth. 
  
\subsection{Calibration}

Finally, we tested incorporating uncertainty quantification with Bayesian model architectures. Using a threshold on the prediction could increase the overall precision of the model, which might be helpful in a scenario where a higher precision is worth the trade-off of a lower recall. The calibration methods used in this paper are from Hagen et al \cite{hagen2022dbcal}. We first used the correct predictions of the validation images to create density trees for both the correct pixels and all pixels. The uncertainty in the test data for each pixel was then calculated as the density of correct pixels over the density of all pixels \cite{hagen2022dbcal}. The Bayesian neural network has dropout layers in the neural network so different inferences (trials) of the model will result in different predictions. These different predictions can be combined with geometric means when calculating the model's uncertainty \cite{hagen2022dbcal}.

For our density estimator, we used a k-Nearest Neighbors density estimator to build k-d trees. The function utilized the KDTrees function from the Python library scikit-learn \cite{scikit-learn}. K-d trees were built for each class using a validation dataset. These k-d trees were then used to calculate the uncertainty for each pixel hypothesis in the test dataset. We used 10 trials for calibration for Bayesian models, and the probability across trails was combined with geometric means, as this has been shown to be a high performing way to average probabilities over multiple independent and identically distributed  trials \cite{nelson2017assessing}. We only used one trial with the Segnet model as the trials for the model, which does not have random dropout layers, were identical. The calibration prediction is calculated by using the class that is most frequently predicted across trials per pixel.

\subsection{General Training}
The models were trained with Python using PyTorch with a dual Nvidia P100 PCI-e based GPU node. The optimizers were the RMSprop and Adam optimizer from the PyTorch package with 1e-4 learn rate that was reduced by a 1e-5 factor after 50 epochs with no improvement in the validation accuracy. The initial weights were either expert-defined weights (EWCE) or the proportion of each class from the training data (WCE). 

We tested every combination of 1) the model architectures: SegNet, SegNeSt, Bayesian SegNet (BayesSegNet), and Small Bayesian SegNet (BayesSegNet), 2) the losses: expert-defined weights (EWCE) and the proportion of each class from the training data (WCE) and 3) optimizers: RMSprop (RMS) and Adam. After training with these combinations, we modified the best combinations to incorporate meta-data, topological loss, focal loss, and uncertainity quantification.

The models were trained with 500 epochs, with the validation accuracy calculated every 10 epochs beginning with the first epoch. The model with the best validation accuracy was saved. Performance metrics were measured with the test subsection of the images.

\subsection{Microstructures}

For each trained model, we calculated the performance metrics for both pixel-level class identification and for microstructure characteristics. The first step for microstructural analysis was to cluster the individual defect pixels into defects. As previously described in Pazdernik et al. \cite{ pazdernik2020microstructural}, pixels within the same class were clustered together using dilation. The defect pixels were dilated into clusters with the function “dilate” from the OpenCV Python Package \cite{opencv_library}. 

For each defect, we recorded the spatial coordination, the bounding box, the area, and if the defect was crossed by a boundary. The boundary indicator was calculated using the boundary extensions created with probabilistic Hough transformations with ``HoughLinesP'' from the OpenCV Python Package and used to identify defect clusters that were intersected by the extended lines \cite{opencv_library}. Additional detail is provided in Pazdernik et al. \cite{opencv_library} \cite{ pazdernik2020microstructural}. 

We calculated several microstructural performance metrics including the intersect over union (IoU) for each defect class, the spatial clustering and dispersion of each defect relative to other defects of the same class and to other defect classes, and the average areas and defect number within each defect class. 

We also calculated whether the proportion of defects on the grain boundary for each defect class was significantly different than 50\%, with a proportion lower than 50\% indicating a higher likelihood of the defect being located within the grain and a proportion above 50\% indicating a higher likelihood of the defect being located on the grain boundary. 

The IoU (intersect over union) per defect class was calculated by matching each expert-labeled defect with the closest predicted defect (the recall Box IoU) and, vice versa, the matching each predicted defect with the closest expert-labeled defect (the Precision Box IoU) and then calculating the intersection of both matched defects' bounding boxes. 

The spatial cluster and dispersion of the defects was calculated with a modified Ripley's K value \cite{ripley1977modelling}, where the K value for the defects in the images is compared with the K values for randomly dispersed samples for each radius value. More defects within a radius distance from each defect than would be expected for a randomly dispersed sample indicates clustering, and less than expected indicates dispersion.

Only defects greater than 0.001888 \(\mu m^{2}\) were considered for all the microstructure analysis to avoid using image artifacts rather than defects. We decided to use a lower threshold value than used in Pazdernik at all (0.0087 \(\mu m^{2}\)) after noticing defects in the irradiated images below the original area threshold.

\subsection{Performance Metrics Formulas}

The performance calculations were previously described in Pazdernik et al. \cite{ pazdernik2020microstructural}. Briefly, these are the metrics used:

\subsubsection{Precision, Recall, and \(F_{D}\)}
To calculate precision and recall for class \(i\), we used equations 1 and 2. \(n_{ij}\) denotes the class \(i\) pixels predicted to be class \(j\), \(n_{ji}\) denotes the class \(j\) pixels predicted to be class \(i\), and \(C\) denotes all the classes. 

(1)
\begin{displaymath}
Precision = \frac{n_{ii}}{\sum_{j\epsilon C}^{}n_{ji}}
\end{displaymath}

(2)
\begin{displaymath}
Recall = \frac{n_{ii}}{\sum_{j\epsilon C}^{}n_{ij}}
\end{displaymath}

When calculating the F-scores, we only used \(F_{D}\) as we found using standard F-scores less practical for evaluation of overall performance. The \(F_{D}\) is a custom F-score that uses  \(B=0.5\) for only the grain class and \(B=2\) for other classes. This weighs precision higher for the background class and recall higher for the other classes. \(n_{C}\) denotes the number of classes. \(g\) denotes the grain class.

(3)
\begin{displaymath}
F_{D} =  \frac{(F_{0.5g} + \sum_{i\not\equiv g}^{}F_{2i})}{n_{C}}
\end{displaymath}

\subsubsection{Intersection over Union }
The intersection over union (IoU) was calculated on both a pixel and a defect level. Equation 4 calculates pixel-level IoU. 

(4)
\begin{displaymath}
IoU = \left( \frac{1}{n_{C}} \right)\sum_{i}^{}\frac{n_{ii}}{\sum_{j}^{}n_{ij}+\sum_{j}^{}n_{ji}-n_{ii}}
\end{displaymath}

In addition to calculating the IoU on a pixel level, we also calculated the IoU for the closest predicted defect to the expert-labeled defect (the Recall Box IoU or Box\textsubscript{r}IoU) and the closest expert-labeled defect  to a predicted class (the Precision Box IoU or Box\textsubscript{p}IoU) for each defect class (equation 5 and 6, respectively). \(T_{i}\) denotes the box outlining each expert-labeled (true) defect of class \(i\) and \(E_{i}\) denotes the same but for predicted (estimated) defects. \(C_{ije}\) denotes the center of each estimated defect, with \(i\) as the defect class, \(j\) as the \(jth\) defect, and \(e\) as the estimated defects, while \(C_{ijt}\) denotes the center of each true defect  with \(t\) as the expert-labeled defects (the true defects). \(A_{ijt}\) denotes the area of the true defects and \(A_{ije}\) denotes the area of the predicted defect \(e\). \(k\) denotes the \(kth\) estimated (\(e\)) defect for Box\textsubscript{r}IoU and the \(kth\) true  (\(t\)) defect for Box\textsubscript{p}IoU. We used \(\epsilon\) =  0.001888 \(\mu\)\(m^{2}\).

(5)
\begin{displaymath}
Box_{r} IoU(T_{i}|E_{i}) = \frac{1}{\left| T_{i} \right|}\sum_{j\in T_{i}}^{}\left\{ \frac{A_{ijt}\bigcap{A_{ike}}^{}}{A_{ijt}\bigcup{A_{ike}}^{}} |argmin_{k\in E_{i}}\left\|C_{ijt} - C_{ike}  \right\|^{2}, A_{ijt} > \epsilon , A_{ike} > \epsilon \right\}
\end{displaymath}

(6)
\begin{displaymath}
Box_{p} IoU(E_{i}|T_{i}) = \frac{1}{\left| E_{i} \right|}\sum_{j\in E_{i}}^{}\left\{ \frac{A_{ije}\bigcap{A_{ikt}}^{}}{A_{ije}\bigcup{A_{ikt}}^{}} |argmin_{k\in T_{i}}\left\|C_{ije} - C_{ikt}  \right\|^{2}, A_{ije} > 	\epsilon , A_{ikt} > \epsilon \right\}
\end{displaymath}

The overall metrics were calculated by averaging the metrics across images.

 \subsubsection{Modified Ripley's K}   
The Ripley function indicates if points are dispersed or clustered around one another. The Ripley function was adapted from the Kest function in the SpatStat R package \cite{baddeley2005spatstat} and the RipleysKEstimator function in the Astropy Python package \cite{2022ApJ...935..167A}. The SpatStat package uses equation (7), which uses \(\frac{a}{n(n-1)}\) rather than \(\frac{a}{n(n)}\) as in the original Ripley equation for an unbiased estimator of Ripley's K (\(R_{K}\)) \cite{marcon2009generalizing} \cite{getis1984interaction} \cite{ripley1977modelling}.

(7)
    \begin{displaymath}
    R_{K} = \frac{a}{n(n-1)}\sum_{i}\sum_{j} I(d_{ij}<r)e_{ij} 
    \end{displaymath}
 
In equation 7, \(a\) denotes the area of the image, \(n\) denotes the number of data points within a defect class, \(i\) and \(j\) denotes ordered pairs of all points within a defect class, \(d_{ij}\) denotes the distance between points \(i\) and \(j\), \(I\) denotes the indicator function for whether the distance is less than the radius input, and \(e_{ij}\) denotes the Ripley correction for the ordered pair \(i\) and \(j\). 
 
For calculating the Ripley values for each ordered pair of defect classes, we modified the Python function to replicate the SpatStat package results for multivariate data by dividing the area by \(n_{1}*n_{2}\) as illustrated  in equation 12 of Dixon et al. \cite{dixon2012ripley}.  This results in the equation 8, where \(i\) denotes the points  and \(n_{i}\) denotes the number of data points in the first defect class and \(j\) denotes the points and \(n_{j}\) denotes the number of data points in the second defect class:

(8)
    \begin{displaymath}
    R_{K}  = \frac{a}{n_{i}*n_{j}}\sum_{i}\sum_{j} I(d_{ij}<r)e_{ij} 
    \end{displaymath}

We also applied the H transformation (the modified Ripley's K or Ripley's H (\(R_{H}\))) to the Ripley result to normalize the data, where r denotes the radius and \(R_{K}\) denotes the output from equation 7 or 8:

 (9)
    \begin{displaymath}
      R_{H}  = \frac{ \sqrt[2]{ R_{K} }}{ \pi } - r
    \end{displaymath}
    
In addition to calculating the Ripley values for predicted and ground truth defects, we calculated the \(99^{th}\) quantile values for the Ripley values of points randomly distributed 1000 times, where the number of points is the same as the number of defects within each class. 

\section{Results}

\subsection{Performance Metrics}
 
We calculated the overall score using the average of \(F_{D}\), the Intersect over Union (IOU), and the average Box IoU score. We found that model with the highest metrics for unirradiated pellet images was the model trained with the SegNet architecture, Adam optimizer, and EWCE loss (\(F_{D}\): 86.1\%, IoU: 75.1\%, average Box IoU: 60.3\%  and average overall score: 73.8\%) (Table \ref{tab:table_performance_metrics_unirradiated}). For irradiated pellet images, the best model was trained with the Small Bayesian SegNet  architecture, Adam optimizer and EWCE loss (\(F_{D}\): 70.5\%, IoU: 59.3\%, average Box IoU: 57.5\% and overall score: 62.5\% (Table \ref{tab:table_performance_metrics_irradiated}). The recall, precision, \(F_{D}\), and IoU were calculated using unweighted averages for each of the classes (grains, boundaries, precipitates, voids and impurities). We used the \(F_{D}\) score as part of our overall metric score because it is calculated using both recall and precision. Appendix Table  \ref{tab:table_performance_metrics_unirradiated_weighted} and Appendix Table  \ref{tab:table_performance_metrics_irradiated_weighted} show the metrics when each class is weighted by the class prevalence within each image, for unirradiated and irradiated pellet images, respectively. 
 
The best models, as measured by the overall score, for both irradiated and unirradiated pellet images both used an Adam optimizer and EWCE loss. With EWCE, the pixels were weighed so a mislabeled grain was valued 10 times less than any other mislabeling in the cross-entropy function. As both Table \ref{tab:recall_precision_irradiated} and Table \ref{tab:recall_precision_unirradiated} show, despite the lower weight of grains, the average recall and precision for the grain class is higher for both unirradiated and irradiated pellet images than the other classes. The boundary class is consistently has both the lowest recall and precision. 

The best model architecture for unirradiated data, the SegNet model (Table \ref{tab:table_performance_metrics_unirradiated}), was also the best model in the Pazdernik et al.\cite{ pazdernik2020microstructural} for predicting pixel classification for unirradiated pellet images. The best model for irradiated data used the Small BayesSegNet architecture (Table \ref{tab:table_performance_metrics_irradiated}). Possibly, the Small BayesSegNet model's high overall score is due to the dropout layers preventing over-fitting of the model. The differences between the three Bayesian models was small (ranging from 62.0\% - 62.5\%) and could be due to the random dropout layers during testing. We decided to use the Small BayesSegNet model as the ``best irradiated model'' to test additional adjustments as it had the highest performance metrics for this trial, and its smaller architecture required a lower computational costs than the other Bayesian architectures for a similar performance. However, all models with the same expert-defined weights seems to perform comparably no matter the architecture. Although we chose one model to compare with the expert-labeled results, we likely could have used any of the other models with expert-defined weights for similar results. 

The performance metrics for irradiated images were generally lower than for unirradiated images. The \(F_{D}\) score is also more consistent across the models for the unirradiated than the irradiated images (Table \ref{tab:table_performance_metrics_irradiated} and (Table \ref{tab:table_performance_metrics_unirradiated}). This is likely because the imaging conditions for the unirradiated pellets is consistent across images, so all models, even those with sub-optimal parameters, are able to achieve a consistently high \(F_{D}\) score. 

The overall performance metric average is lowered by including the box IoU scores in the average. This metric is important because the defect location is an quantification measure of pellets. The metrics involves multiple steps – segmenting the images into each defects class correctly with multi-class segmentation, matching each defect cluster in the predicted and expert-labeled images, then calculating the bounding box average for each match. Any defect without an overlapping match will have zero overlap and will lower the average Box IoU score. Considering the difficulty of obtaining a high Box IoU score, the best models had an high overall score. This indicates that utilizing these models could be an efficient alternative to expert-labeling of images. 

Supplemental Figures \ref{fig:S1_unirradiated_image_1}, \ref{fig:S2_unirradiated_image_2}, \ref{fig:S3_unirradiated_image_3}, \ref{fig:S4_unirradiated_image_4}, \ref{fig:S5_unirradiated_image_5} show the input, predicted, and expert-labeled test images for images 1-5, respectively.

\begingroup
\squeezetable
\begin{table}

\caption {\label{tab:table_performance_metrics_unirradiated}  Performance Metrics for Unirradiated Pellets: Precision, recall, and \(F_{D}\) for pixel metrics, and Box IoU precision (\(Box_{p}\)), Box IoU recall (\(Box_{r}\)), average of Box recall and Precision (\(Box_{a}\)), and the average of the \(F_{D}\), IoU, and Box A scores for unirradiated pellet images. Metrics are for the Bayesian SegNeSt (BayesSegNeSt), Bayesian SegNet (Bayes), SegNet, and Small Bayesian SegNet (SmallBayes) models and using either a EWCE or WCE loss and Adam and RMS optimizer. The models used for the additional adjustments are a SegNet model with a Adam optimizer and Focal (FL) or Topological (Topo) loss. The best score for each metric is in bold.} 

\begin{ruledtabular}
\begin{tabular}{lll|llll|llll}
Models & Opt. & Loss & Precision & Recall & \(F_{D}\) & IoU & \(Box_{p}\) & \(Box_{r}\) & \(Box_{a}\) & Avg  \\    
 \hline
 \multirow{4}{*}{BayesSegNeSt}   &  \multirow{2}{*}{Adam}   & EWCE  & 79.4\% & 89.1\% & \textbf{86.9\%} & 73.1\% & 54.2\% & 61.9\% & 58.1\% & 72.7\%    \\ 
 \cline{3-11} 
 &     & WCE  & 70.5\% & \textbf{90.6\%} & 84.7\% & 65.4\% & 43.8\% & 53.7\% & 48.8\% & 66.3\%      \\  
  \cline{3-11} 
 &   \multirow{2}{*}{RMS}   & EWCE  & 79.5\% & 88.9\% & 86.7\% & 72.8\% & 56.7\% & 62\% & 59.3\% & 72.9\%    \\ 
 \cline{3-11}   
  &     & WCE  & 73\% & 88.7\% & 84.5\% & 66.9\% & 47.9\% & 57\% & 52.4\% & 68\%     \\  
\hline
 \multirow{4}{*}{BayesSegNet}   &  \multirow{2}{*}{Adam}   & EWCE  & 79.1\% & 88.6\% & 86.3\% & 72.3\% & 54.6\% & 63\% & 58.8\% & 72.4\%   \\ 
 \cline{3-11} 
 &     & WCE  & 69.5\% & 88.8\% & 82.9\% & 63.4\% & 43\% & 53.3\% & 48.2\% & 64.8\%   \\  
  \cline{3-11} 
 &   \multirow{2}{*}{RMS}   & EWCE  & 78\% & 87.9\% & 85.5\% & 70.9\% & 52.5\% & 60.8\% & 56.6\% & 71\%   \\ 
 \cline{3-11}   
  &     & WCE  & 70.5\% & 89.2\% & 83.8\% & 64.7\% & 42.6\% & 54.7\% & 48.6\% & 65.7\%  \\  
  \hline

 \multirow{4}{*}{SegNet}   &  \multirow{2}{*}{Adam}   & EWCE & \textbf{83.7}\% & 86.7\% & 86.1\% & \textbf{75.1\%} & 55.9\% & 64.6\% & 60.3\% & \textbf{73.8\%}  \\ 
 \cline{3-11} 
 &     & WCE  & 82.1\% & 87.8\% & 86.5\% & 74.3\% & 56.3\% & 64.3\% & \textbf{60.3\%} & 73.7\% \\  
  \cline{3-11} 
 &   \multirow{2}{*}{RMS}   & EWCE  & 83.5\% & 85.4\% & 85\% & 74.1\% & 54.2\% & \textbf{64.7\%} & 59.5\% & 72.8\%    \\ 
 \cline{3-11}   
  &     & WCE  & 81.5\% & 86.8\% & 85.6\% & 73.2\% & 54.2\% & 63.6\% & 58.9\% & 72.6\%  \\  
  \hline
 \multirow{4}{*}{SmallBayes}   &  \multirow{2}{*}{Adam}   & EWCE & 79\% & 89.2\% & 86.9\% & 72.7\% & \textbf{56.6\%} & 62.5\% & 59.6\% & 73.1\%  \\ 
 \cline{3-11} 
 &     & WCE  & 70.7\% & 89.8\% & 83.9\% & 65\% & 41.3\% & 54.2\% & 47.7\% & 65.6\% \\  
  \cline{3-11} 
 &   \multirow{2}{*}{RMS}   & EWCE  & 79.1\% & 89.3\% & 86.9\% & 72.8\% & 54\% & 62.4\% & 58.2\% & 72.6\%    \\ 
 \cline{3-11}   
  &     & WCE  & 71\% & 89.3\% & 83.9\% & 65.4\% & 41.6\% & 54.6\% & 48.1\% & 65.8\%     \\  
 \hline
 \multirow{2}{*}{SegNet}   &   \multirow{2}{*}{Adam}  & Focal Loss  & 82.5\% & 86.5\% & 85.5\% & 73.8\% & 54.3\% & 62.7\% & 58.5\% & 72.6\% \\ 
  \cline{3-11} 
  &    & Topology  & 83.1\% & 86.1\% & 85.4\% & 74.2\% & 54.2\% & 63.4\% & 58.8\% & 72.8\%  \\ 
 
\end{tabular}
\end{ruledtabular}

\end{table}
\endgroup

\begingroup
\squeezetable
\begin{table}
\caption {\label{tab:table_performance_metrics_irradiated}  Performance Metrics for Irradiated Pellets: Precision, recall, and \(F_{D}\) for pixel metrics, and Box IoU precision (\(Box_{p}\)), Box IoU recall (\(Box_{r}\)), average of Box recall and Precision (\(Box_{a}\)), and the average of the \(F_{D}\), IoU, and Box A scores for irradiated pellet images. Metrics are for the Bayesian SegNeSt (BayesSegNeSt), Bayesian SegNet (Bayes), SegNet, and Small Bayesian SegNet (SmallBayes) models and  using either a Expert Weighted Cross Entropy (EWCE) or Weighted Cross-Entropy (WCE) loss and Adam and RMS optimizer for the initial models. The models used for the additional adjustments are the Small Bayesian SegNet (SmallBayes) model with a Adam optimizer and Focal (FL) or Topological (Topo) loss or a SegNet model with meta-data and a Adam optimizer and EWCE loss. This table does not include the SegNet Meta model as the meta-data was unavailable for these images. The best score for each metric is in bold. The best score for each metric is in bold.} 

\begin{ruledtabular}
\begin{tabular}{lll|llll|llll}
Models & Opt. & Loss & Precision & Recall &  \(F_{D}\) & IoU & \(Box_{p}\) & \(Box_{r}\) & \(Box_{a}\) & Avg \\    
 \hline
 \multirow{4}{*}{BayesSegNeSt}   &  \multirow{2}{*}{Adam}   & EWCE  & 77.1\% & 64.6\% & 79.5\% & 55.5\% & 50.6\% & 51.2\% & 50.9\% & 62\%  \\ 
 \cline{3-11} 
 &     & WCE  & 65.4\% & \textbf{74\%} & 71.5\% & 57.1\% & 50.3\% & 49.2\% & 49.8\% & 59.4\%      \\  
  \cline{3-11} 
 &   \multirow{2}{*}{RMS}   & EWCE  & \textbf{79.7\%} & 56.1\% & 89.5\% & 45.3\% & 46.8\% & 39.6\% & 42.5\% & 59.1\%    \\ 
 \cline{3-11}   
  &     & WCE  & 58.1\% & 67.3\% & 65\% & 51.8\% & 49.8\% & 44.6\% & 47.2\% & 54.7\%  \\  
\hline
 \multirow{4}{*}{Bayes SegNet}   &  \multirow{2}{*}{Adam}   & EWCE  & 67.5\% & 70.8\% & 70\% & 59.2\% & 61.2\% & 52.1\% & 56.7\% & 62\%      \\ 
 \cline{3-11} 
 &     & WCE  & 55.9\% & 69.4\% & 66.1\% & 49.3\% & 38.3\% & 46.3\% & 42.3\% & 52.5\%  \\  
  \cline{3-11} 
 &   \multirow{2}{*}{RMS}   & EWCE  &79\% & 56.1\% & \textbf{89.9\%} & 44.9\% & 45.4\% & 39.2\% & 41.7\% & 58.8\%      \\ 
 \cline{3-11}   
  &     & WCE  & 59.3\% & 70\% & 64\% & 47.9\% & 38.9\% & 47.6\% & 43.3\% & 51.7\%    \\  
  \hline
 \multirow{4}{*}{SegNet}   &  \multirow{2}{*}{Adam}   & EWCE  & 74.5\% & 67.7\% & 67.9\% & 59.2\% & 55.8\% & \textbf{54.3\%} & 55\% & 60.7\%    \\ 
 \cline{3-11} 
 &     & WCE  & 68.4\% & 71.6\% & 69.9\% & 56.6\% & 49\% & 48.9\% & 48.9\% & 58.5\% \\  
  \cline{3-11} 
 &   \multirow{2}{*}{RMS}   & EWCE  & 72.9\% & 68.7\% & 68.7\% & 59.5\% & 53.7\% & 50.9\% & 52.3\% & 60.2\%  \\ 
 \cline{3-11}   
  &     & WCE  & 69.2\% & 67\% & 67.1\% & 58.2\% & 50.4\% & 52\% & 51.2\% & 58.8\%   \\  
  \hline
 \multirow{4}{*}{Small Bayes}   &  \multirow{2}{*}{Adam}   & EWCE  & 71.8\% & 71.6\% & 70.5\% & 59.3\% & \textbf{63.3\%} & 51.8\% & \textbf{57.5\%} & \textbf{62.5\%}   \\ 
 \cline{3-11} 
 &     & WCE  & 61.4\% & 73.9\% & 69.6\% & 54.5\% & 45.9\% & 49.4\% & 47.7\% & 57.3\%     \\  
  \cline{3-11} 
 &   \multirow{2}{*}{RMS}   & EWCE  & 67.3\% & 72\% & 70.8\% & 59\% & 50.6\% & 52.1\% & 51.4\% & 60.4\%  \\ 
 \cline{3-11}   
  &     & WCE  & 60.8\% & 71.1\% & 67.1\% & 52.8\% & 43.5\% & 48.3\% & 45.9\% & 55.3\%    \\  
 \hline
 \multirow{2}{*}{SmallBayes} & \multirow{2}{*}{Adam} & Focal Loss  & 70.3\% & 56\% & 75.3\% & 45.3\% & 47.1\% & 39.7\% & 42.7\% & 54.4\% \\ 

 & & Topology  & 70.1\% & 71.3\% & 70.2\% & 58.4\% & 53.8\% & 50.8\% & 52.3\% & 60.3\%  \\ 

 SegNet Meta &  Adam  & EWCE  & 75.6\% & 69\% & 69.1\% & \textbf{60.2\%} & 58.2\% & 54.3\% & 56.3\% & 61.8\% \\ 
\end{tabular}
\end{ruledtabular}
\end{table}
\endgroup

\begingroup
\squeezetable
\begin{table}

\caption {\label{tab:recall_precision_unirradiated} Unirradiated Precision and Recall: The precision and recall of unirradiated  pellet images for the best unirradiated model: SegNet with EWCE loss and Adam optimizers.} 
    \begin{ruledtabular}
    \begin{tabular}{|l|l|l|}
    \hline
        \textbf{Category} & \textbf{Precision} & \textbf{Recall} \\ \hline
        Grain & 98.5\% & 97.9\% \\ \hline
        Boundary & 73.1\% & 78.3\% \\ \hline
        Void & 80.7\% & 83.0\% \\ \hline
        Impurity & 75.4\% & 82.3\% \\ \hline
        Precipitate & 90.6\% & 92.1\% \\ \hline

\end{tabular}
\end{ruledtabular}
\end{table}
\endgroup

\begingroup
\squeezetable
\begin{table}
\caption {
\label{tab:recall_precision_irradiated} Irradiated Precision and Recall:  Precision and recall of irradiated  pellet images for the best irradiated model: Small Bayes SegNet with EWCE loss and Adam optimizers.}
    \begin{ruledtabular}
    \begin{tabular}{|l|l|l|}
    \hline
        \textbf{Category} & \textbf{Precision} & \textbf{Recall} \\ \hline
        Grain & 97.1\% & 94.3\% \\ \hline
        Boundary & 23.2\% & 38.2\% \\ \hline
        Void & 83.4\% & 89.9\% \\ \hline
        Impurity & 79.8\% & 44.0\% \\ \hline
        Precipitate & 75.5\% & 91.6\% \\ \hline

\end{tabular}
\end{ruledtabular}
\end{table}
\endgroup

\subsection{Performance Metrics for Additional Adjustments}
 
Utilizing focal loss lowered performance of the model (Table \ref{tab:table_performance_metrics_unirradiated}) and (Table \ref{tab:table_performance_metrics_irradiated}). For the irradiated images, there was a large difference in the average overall score between the model with and without Focal Loss (54.4\% vs 62.5\%, respectively), and the score was also lower for unirradiated images. Perhaps the weighted cross-entropy already weighs the non-grain classes losses heavily enough and including the focal loss ultimately over-weighs defect losses.

Topological loss also did not improve the performance metrics of the irradiated or unirradiated model contrary to our hypothesis (Table \ref{tab:table_performance_metrics_unirradiated}). It is possible that incorporating topological loss did not increase the performance metrics because the 128-pixel chips of the training images broke up the boundary paths and thereby broke up the topology.

Using the SegNet architecture with meta-data also did not improve the overall score compared with the best irradiated model (Small BayesSegNet with EWCE loss and Adam optimizer) (Table \ref{tab:table_performance_metrics_unirradiated}). Perhaps the meta-data did not provide useful information for the model. 

In addition to improving the performance metrics of the models, we hoped that these additional adjustments to the model (focal loss, topological loss, and adding in meta-data) would increase the boundary precision and recall. For the unirradiated images, incorporating focal loss decreased both the boundary precision and recall. The model with the topological loss performed better that the best model for unirradiated pellets for boundary precision only (Boundary precision: 72.8\% and 73.9\% for focal loss and topological loss, respectively vs 73.1\% for the best unirradiated model, and boundary recall: 77.8\% and 77.8\% for focal loss and  topological loss, respectively vs 78.4\% for the best  model for unirradiated data) (Table \ref{tab:boundary_metrics}). However, the slightly higher boundary precision for the topological model does not actually indicate an improvement due to incorporating topological loss. The model architecture saved the model from the epoch with the best validation accuracy, and that occurred before the topological loss was incorporated at 475 epochs. Any difference with the normal loss is due to small variations in training runs, which may be larger than usual because the random dropout-layers in Bayesian models introduced randomness during both training and inference of the test data. 

\begin{table}
\caption {
\label{tab:boundary_metrics} Boundaries:  Precision and recall of grain boundaries in irradiated and unirradiated pellet images.}
    \begin{ruledtabular}
\begin{tabular}{|l|l|l|l|l|l|}
 Image & Models & Opt. & Loss & Precision Boundaries & Recall Boundaries \\   
 \hline
   \multirow{3}{*}{Unirradiated} & \multirow{3}{*}{SegNet} & \multirow{3}{*}{Adam} & EWCE  & 73.1\% & 78.4\%  \\ 
    \cline{4-6} 
 & & & FL & 72.8\% & 77.8\%  \\ 
    \cline{4-6} 
 & & & Topology  & 73.9\% & 77.8\%  \\ 
 \hline
    \multirow{4}{*}{Irradiated} & \multirow{3}{*}{SmallBayes} & \multirow{3}{*}{Adam} & EWCE & 23.2\% & 38.2\%  \\ 
    \cline{4-6} 
   &  & & Focal Loss  & 20.3\% & 0.1\%  \\ 
    \cline{4-6} 
    & & & Topology  & 24.9\% & 39.1\% \\ 
    \cline{2-6} 
  & Segnet Meta & Adam & EWCE  & 34\% & 23.9\%  \\ 
\end{tabular}
\end{ruledtabular}
\end{table}

There was also not a consistent improvement for the irradiated boundary metrics utilizing focal loss, topological loss or incorporating meta-data. Incorporating meta-data increased the boundary precision, but the recall decreased by more than the precision increased, and the higher precision for the topology model is not due to the incorporated topological loss. (Boundary precision: 20.3\%, 24.9\%,  and 34.0\% for focal loss, topological loss, and meta-data, respectively, vs. 23.2\% for the best irradiated model, and boundary recall: 0.1\%, 39.1\%, and 23.9 \% for focal loss and topological loss, respectively vs 38.2\% for the best irradiation model (Table \ref{tab:boundary_metrics}).)

\subsection{Calibration Uncertainity}

\begin{figure}[htbp]
    
    \includegraphics[width=\textwidth]{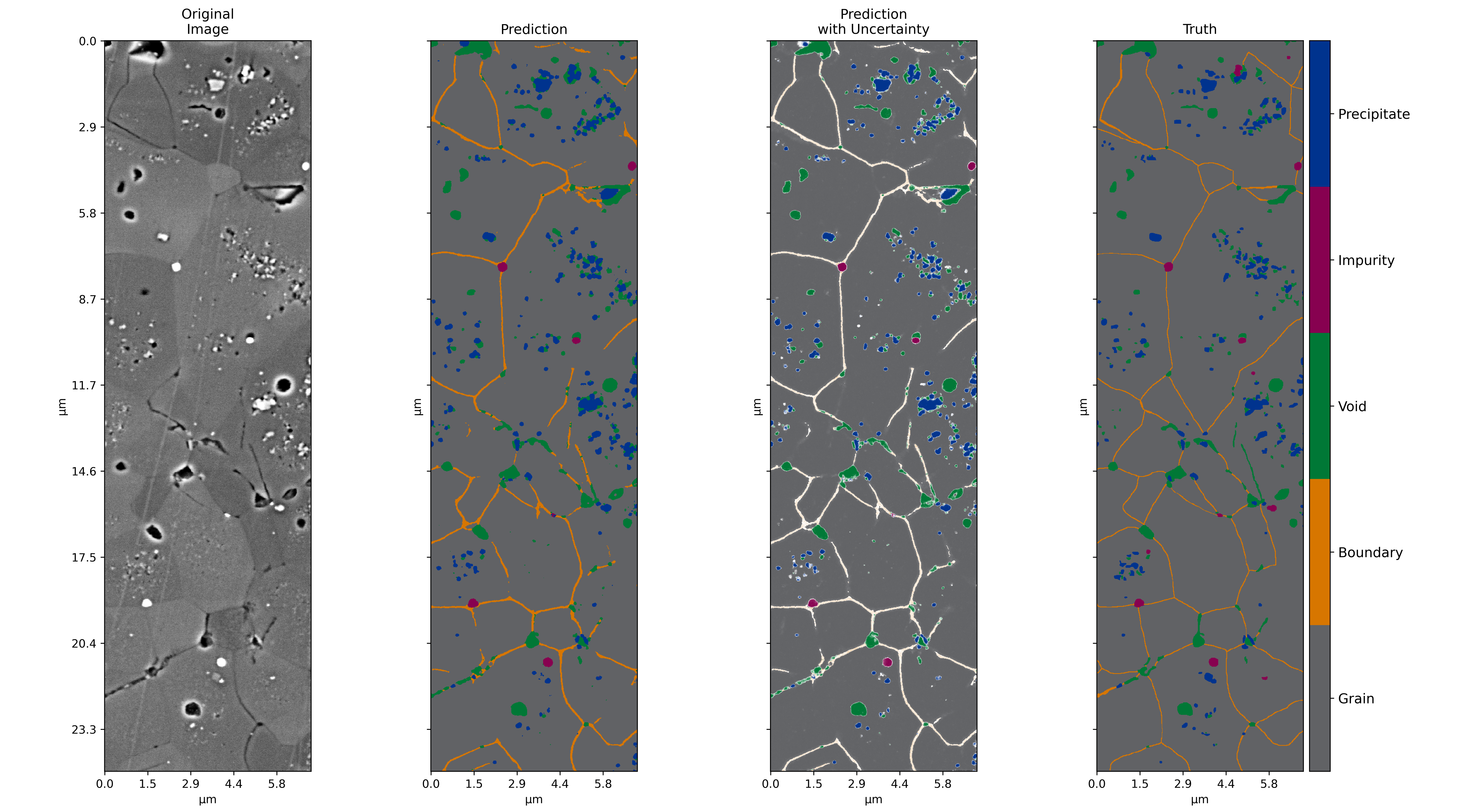}
      \caption{\label{fig:calibration_images} Predictions: From left to right, 1) the original test section of ``Image 4'', 2) the calibration predicted image, created with multiple trial aggregations, 3) the predicted image with uncertainty, with the percentage of certainty represented by the alpha value of the pixels (high opacity indicates a high percentage of uncertainty), and 4) the expert-labeled image. The predicted images were created using the results of multiple trials with the best irradiated model, which used these parameters: Small Bayesian SegNet model architecture, EWCE loss, and Adam optimizer.}
    
\end{figure}

When using the 95\% calibration uncertainty threshold for the irradiated pellet images, with the Small BayesSegNet model with Adam optimizer and EWCE loss model, the general recall was much lower compared to using a single trial of the model without a uncertainty threshold (47.2\% vs 71.6\%, respectively), but the precision was higher (97.2\% vs 71.8\%). For the unirradiated best model both recall and precision were lower (78.9\% vs 83.7\% precision and 55.2\% vs 86.7\% recall, with and without the calibration threshold). Figure \ref{fig:calibration_images} shows the calibration uncertainty percentage as the alpha value of the pixel values for irradiated Image 4. Using the calibration uncertainty threshold does increase the precision with repeated trials, and therefore might be useful in situations where precision is more important than recall. 

Aggregating multiple trials for the Small Bayesian SegNet model for the calibration prediction led to a similiar overall metric compared to using a single trial (61.5\% with multiple trials vs 62.5\% in a single trial).  In Figure \ref{fig:calibration_images}, the prediction using the best irradiation model and expert-labeled images (labeled ``truth'') can be compared visually. Although the two images are similar, it is evident the predicted image's boundaries is generally thicker than the expert-labeled images and fragmented in places.

\subsection{Pixel Proportion}

\begin{figure}[htbp]
    
    \includegraphics[width=\textwidth]{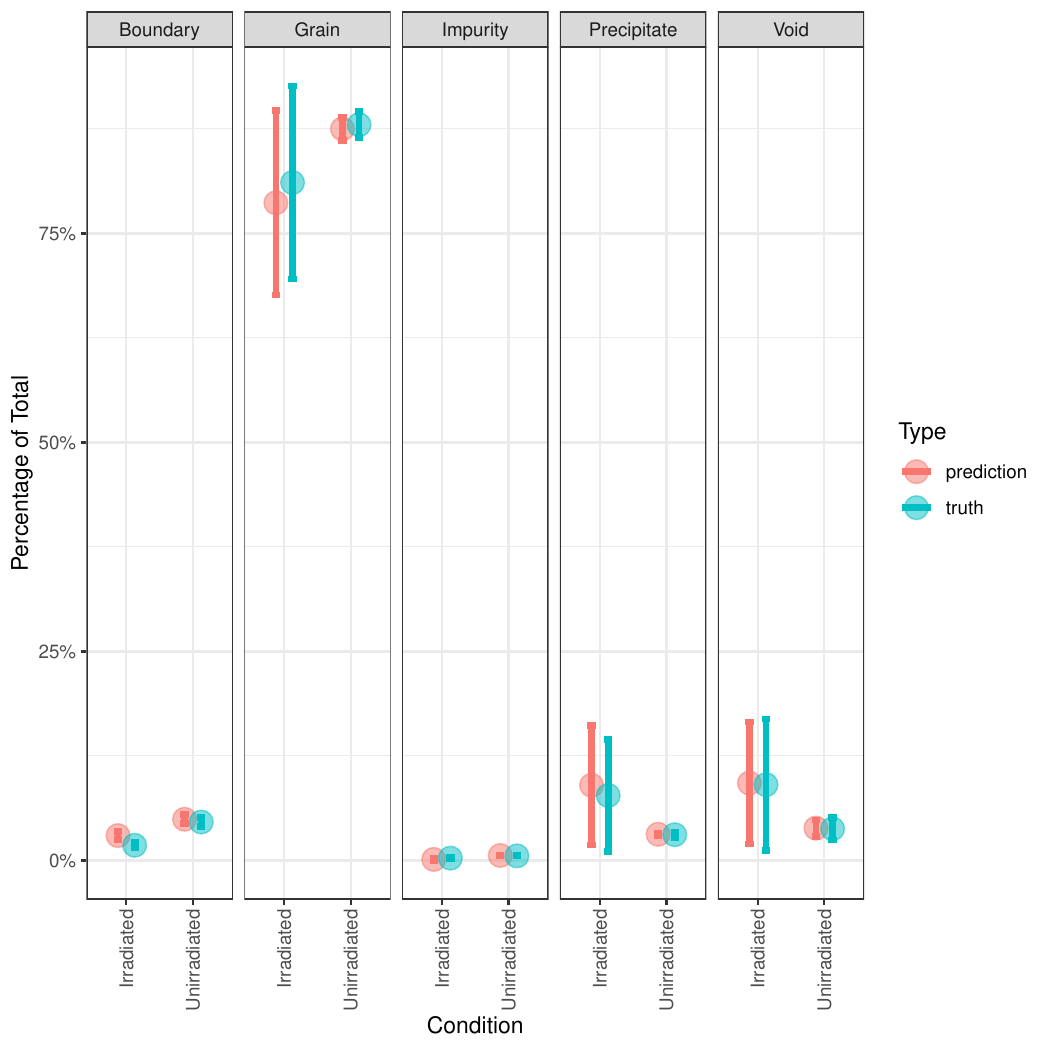}
    \caption{\label{fig:proportion_plot} Pixel Proportion: The mean proportion of defects in the irradiated and unirradiated images, with the error bars representing the 95\% confidence interval.}
    
\end{figure}

Figure \ref{fig:proportion_plot} shows the 95\% confidence interval for the percentage of the pixels classified as either impurities, voids, or precipitates. The confidence interval is calculated as the standard error for each of the three images for the irradiated images and two images for the unirradiated images, multiplied by the \(97.5^{th}\) percentile point of the standard normal distribution. The expert-labeled and predicted image proportion averages are similar, which is promising for using predicted images rather than expert-labeled segmentation for pixel proportion metrics. 

For both the expert-labeled and predicted segmentation, the irradiated pellet images had a high average proportion of precipitates and voids. However, the confidence interval is large, which indicates that while a high proportion of precipitates and voids is a possible characteristic of irradiated pellets, it is not a consistent characteristic across images. 

\subsection{Defect Area}

\begin{figure}[htbp]
    
    \includegraphics[width=\textwidth]{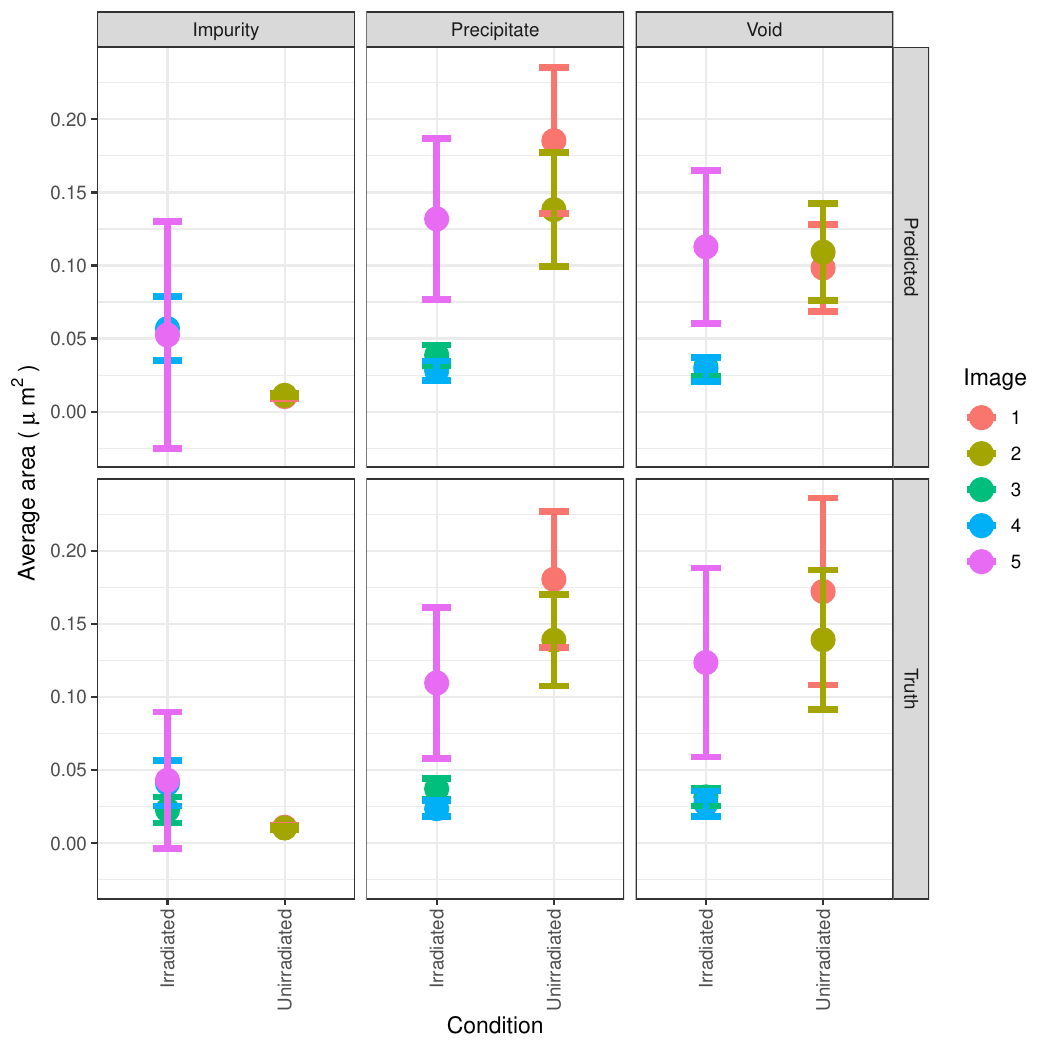}
    \caption{\label{fig:Truth_Defect_Area} Defect Average: The average area of each defects for irradiated and unirradiated pellet images. The error bar represents the 95\% confidence interval.}
    
\end{figure}

The variance in both the defect areas and the confidence interval for irradiated images makes it difficult to identify any consistent difference in defect area between irradiated and unirradiated images  (Figure \ref{fig:Truth_Defect_Area}). (Appendix Table  \ref{tab:table_average_area_density}). The confidence value range for the unirradiated images are within the confidence value range of values for irradiated images. 

As with the pixel proportions, predicted images have the same pattern and similar values for defect areas as the expert-labeled images, which again indicates the predicted images is a valid alternative for hand-labeling when quantifying the areas of the defects.

The generally lower performance metrics across all models for the irradiated pellet images (Table \ref{tab:table_performance_metrics_irradiated}) compared to unirradiated pellet images might reflect the generally smaller size of the defects in the irradiated compared to unirradiated pellet images. The smaller size could make the defects more difficult for the models to distinguish from grain artifacts.

\subsection{Defect Density}

\begin{figure}[htbp]
    
    \includegraphics[width=\textwidth]{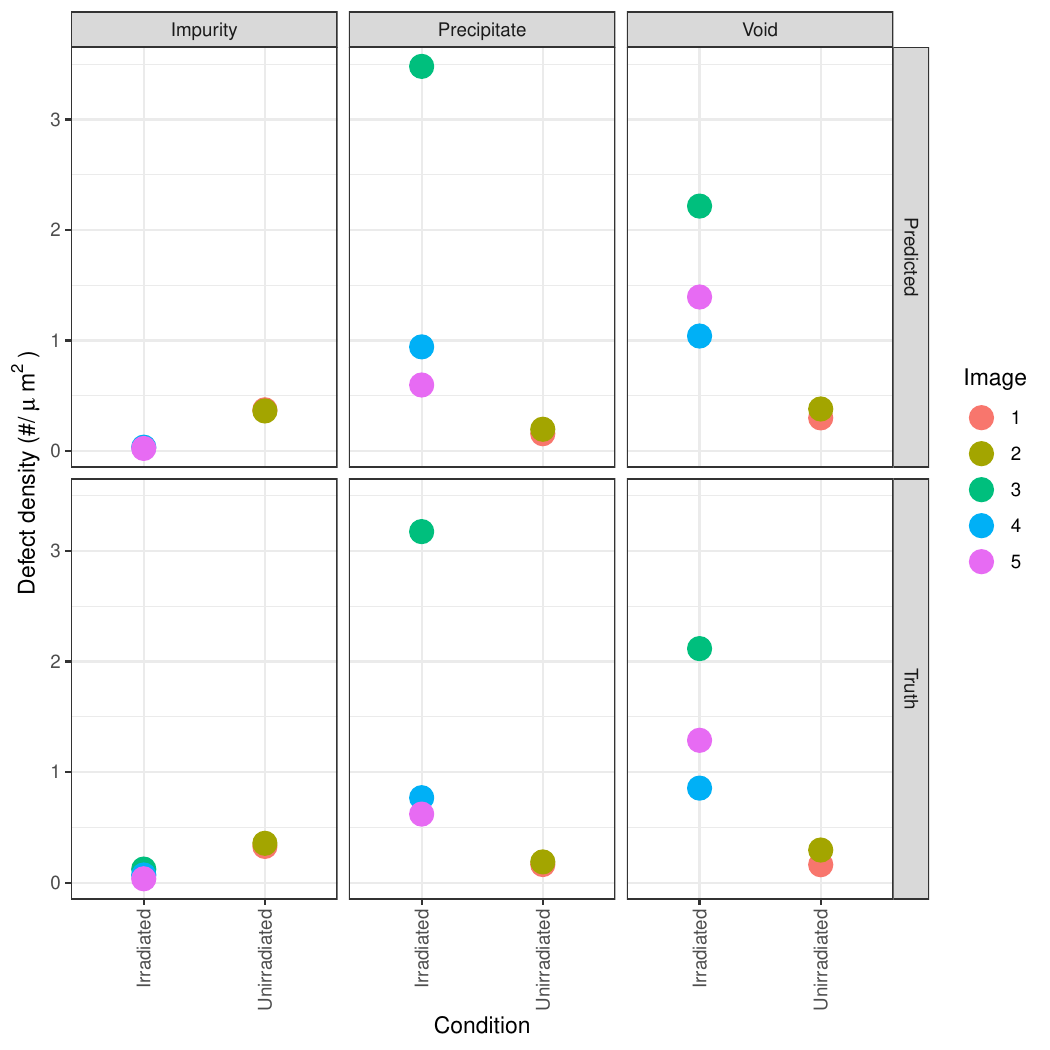}
    \caption{\label{fig: Defect_Density} Defect Density: The density of defects per image, with density calculated as the number of defects over the test image area.}
    
\end{figure}

The density of precipitates and voids is higher for all images of irradiated pellets than  unirradiated pellet images, while the density of impurities is lower (Appendix Table  \ref{tab:table_average_area_density} and Figure \ref{fig: Defect_Density}). The precipitate densities are above 0.50 \(\#/\mu m^{2}\) and the void densities are above 0.85 \(\#/\mu m^{2}\) for both predicted and expert labeled-images irradiated images, while the density of precipitates is below .20 \(\#/\mu m^{2}\) and density of voids is below .40 \(\#/\mu m^{2}\)  for both predicted and expert-labeled unirradiated pellet images. For both the predicted and expert-labeled images, there is a distinct difference between the irradiated and unirradiated image pellets for densities. There are no confidence intervals as each dot in the Figure \ref{fig: Defect_Density} represents just one image. 

\subsection{Defects on Boundaries}

\begin{figure}[htbp]
    
    \includegraphics[width=\textwidth]{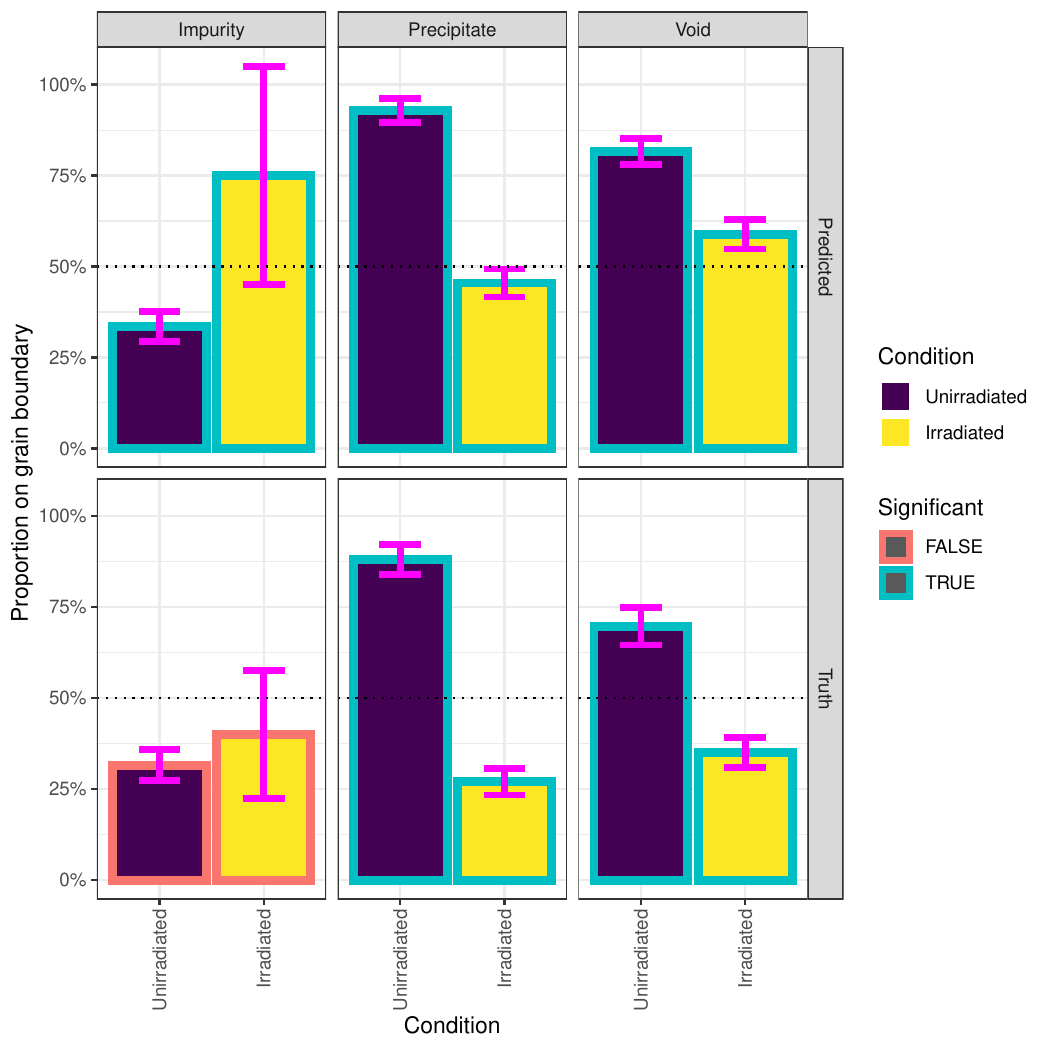}
    \caption{\label{fig:prop_grain_boundary} Proportion on Grain Boundary: Percentage of each defect type on the grain boundary. The error bar represents the 95\% confidence interval of the proportions. The bar outline color represents if there is a significance difference from 50\% in the proportion of defects on the boundary.}
    
\end{figure}

There is a consistent difference between irradiated and unirradiated pellet images in regards to the proportion of defects on the boundaries, which is lower for precipitates and voids in irradiated images (Figure \ref{fig:prop_grain_boundary} and Appendix Table  \ref{tab:prop_grain_boundary_table} )

Both the predicted and expert-labeled images had a proportion significantly greater than 50\% for precipitates and voids for unirradiated images and significantly less than 50\% for precipitates for irradiated pellet images as measured by a one-sample test of equal proportions. However, there was a difference between the predicted and expert-labeled images in whether the proportion of defects was significantly below 50\% for for impurities and if the proportion of defects was above or below 50\% for voids and impurities with irradiated pellet images. The boundary class consistently has the lowest recall and precision of all the pixel classes, which could explain why the proportion of voids and impurity defects on the boundary might not be accurately predicted with a DCNN.

\subsection{Co-clustering of Defects: Univariate and Multivariate}

\begin{figure}[b]
    
    \includegraphics[width=\textwidth]{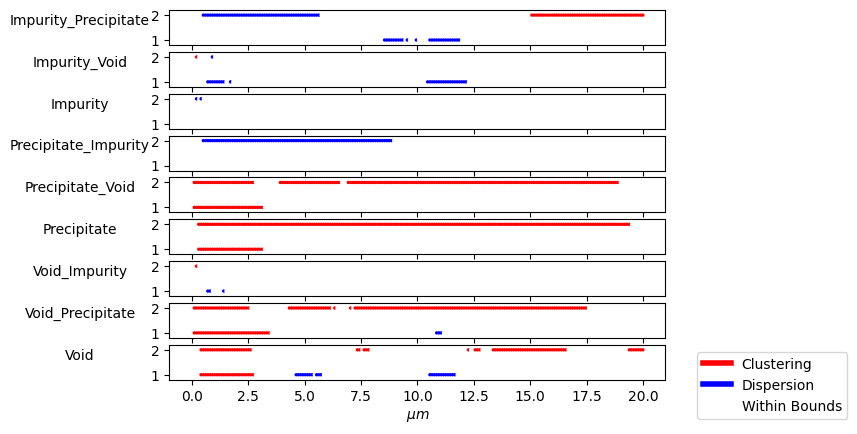}
    \caption{\label{fig:defect_ripleys_unirradiated_predicted} Ripley Results for the Unirradiated Predicted Image: The numbers in the y axis represent the individual images and the scatter dots in the graph represent clustering (red) and dispersion (blue) of defects of the image at the radius location (indicated in the x axis). If  there  are  two  defects separated by an underscore in ``Combination'', the second defect is clustering around the first defect.}
    
\end{figure}

\begin{figure}[htbp]
    
    \includegraphics[width=\textwidth]{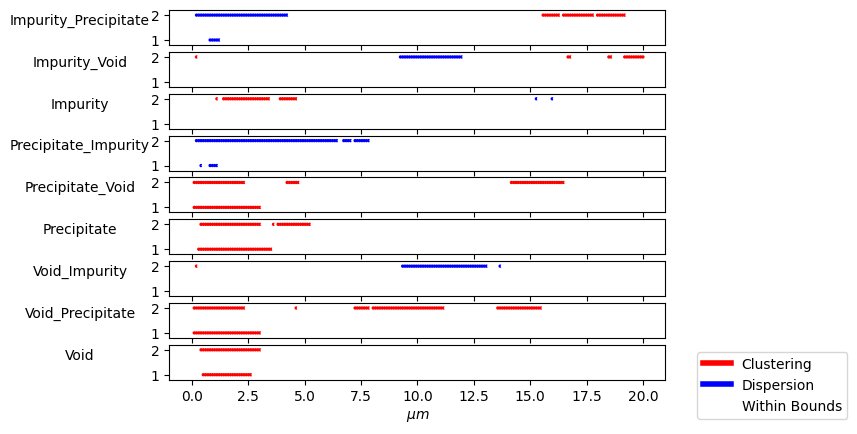}
    \caption{\label{fig:defect_ripleys_Expert-Labeled_unirradiated} Ripley Results for the Unirradiated Expert-Labeled Image: The numbers in the y axis represent the individual images and the scatter dots in the graph represent clustering (red) and dispersion (blue) of defects of the image at the radius location (indicated in the x axis). The first defect type in a combination, such as Impurity\_Precipitate, is the defect the second defect is clustering around.}
    
\end{figure}
\begin{figure}[htbp]
    
    \includegraphics[width=\textwidth]{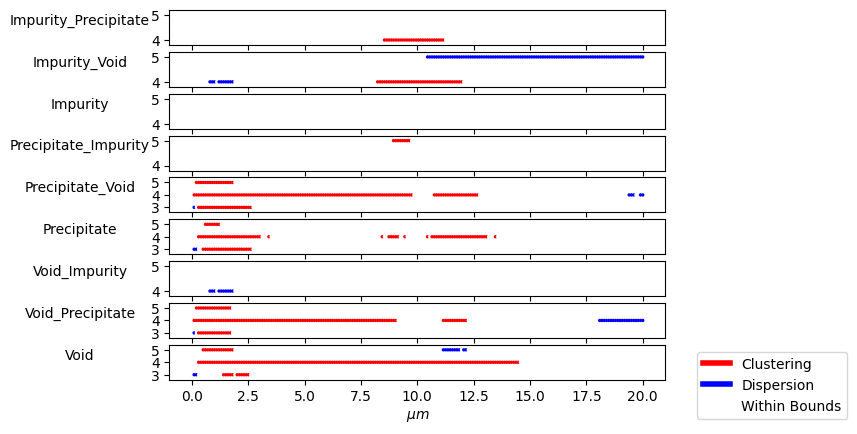}
    \caption{\label{fig:defect_ripleys_predicted_irradiated} Ripley Results for the Irradiated Predicted Image: The numbers in the y axis represent the individual images and the scatter dots in the graph represent clustering (red) and dispersion (blue) of defects of the image at the radius location (indicated in the x axis). If  there  are  two  defects separated by an underscore in ``Combination'', the second defect is clustering around the first defect. There are no impurities over the area threshold for Image 3 so impurities for Image 3 are not depicted in this graph.}
    
\end{figure}

\begin{figure}[htbp]
    
    \includegraphics[width=\textwidth]{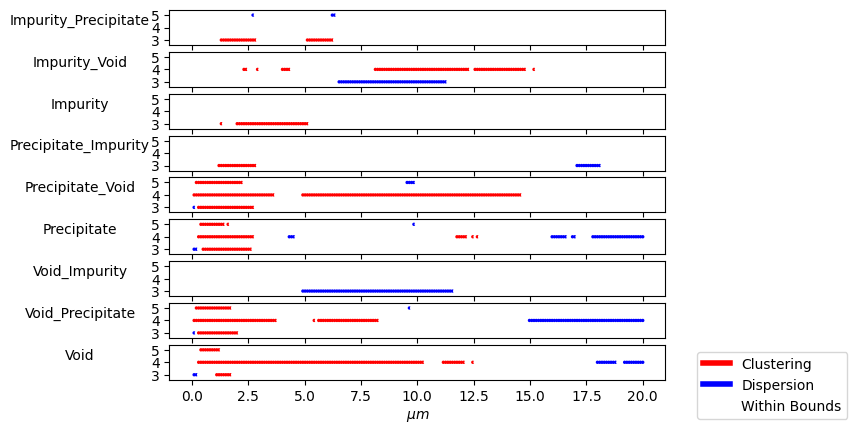}
    \caption{\label{fig:defect_ripleys_Expert-Labeled_irradiated} Ripley Results for the Irradiated Expert-Labeled Image: The numbers in the y axis represent the individual images and the scatter dots in the graph represent clustering (red) and dispersion (blue) of defects of the image at the radius location (indicated in the x axis). If  there  are  two  defects separated by an underscore in ``Combination'', the second defect is clustering around the first defect.}
    
\end{figure}

The radius values where the Ripley values of the defects was either above or below the \(99^{th}\) quantile of the random distribution are shown in Figure \ref{fig:defect_ripleys_unirradiated_predicted} and Figure \ref{fig:defect_ripleys_Expert-Labeled_unirradiated}, the predicted and expert-labeled unirradiated pellet images, respectively and in Figure \ref{fig:defect_ripleys_predicted_irradiated} and Figure \ref{fig:defect_ripleys_Expert-Labeled_irradiated}, with the predicted and expert-labeled irradiated pellet images, respectively. For both the predicted and the expert-labeled irradiated and unirradiated pellet image, the precipitates were clustered around the voids, the voids around the precipitates and the voids and precipitates were clustered with themselves between about 0.0 and 2.5 \(\mu m\) for all five images. 

Impurities and other combinations of defects do not have consistent clustering or dispersion patterns across either unirradiated or irradiated pellet images.There are also a number of mismatches in the clustering found in the expert-labeled images vs the predicted images, especially for clustering involving impurities for irradiated pellet images and clustering above 5 \(\mu m\) for unirradiated pellet images for precipitates and voids clustering around precipitates. Impurities are in such a low number that any differences in impurity labeling seems likely to have a large effect on the Ripley results. It's possible that dispersion and clustering for voids and precipitates do not seem to be accurately predicted at larger radiuses because of the test image's width, which is less than 7 \(\mu m\) for the irradiated pellet images. This makes it more likely that a defect's radius will extend beyond the image, requiring a correction in the Ripley function to adjust for the missing part of the radius's area. For this reason, the Ripley values for larger radiuses will be calculated from less data and might be more sensitive to small differences in defect distribution. 

The radius range with continuous clustering is more consistent between the unirradiated images than for the irradiated images. For example, in regards to voids clustering around precipitates (``precipitate\_voids'') in predicted images, the irradiated pellets has clustering ending before 9.85 \(\mu m\) in ``Image 4'' and ending before 1.9 \(\mu m\) in ``Image 5'' (Figure \ref{fig:defect_ripleys_predicted_irradiated}) where as the unirradiated pellets have clustering ending before 3.12 \(\mu m\) (``Image 1'') and ending before 2.41 \(\mu m\) (``Image 2'') (Figure \ref{fig:defect_ripleys_unirradiated_predicted}). The increased variability of Ripley clustering in the irradiated pellet images is consistent with the variability of other microstructure quantification's in irradiated pellet images, such as defect density, average area of defects, and defect proportions of total pixels. 

\section{Discussion}

We have confirmed that, like with the unirradiated image analyzed in Pazdernik et al., DCNN can successfully segment irradiated images, and this segmentation can be processed into quantifiable microstructural features \cite{ pazdernik2020microstructural}. Considering that expert-labeling can take up to 100 hours per image, the DCNN makes structural analysis of pellets during the irradiation process much more feasible. The DCNN segmentation could be the first step in tracking microstructural evolution during irradiation and predicting pellet performance. 

Although we found that SegNet was the best model for unirradiated images and Small Bayesian SegNet was the best model for irradiated images, there was only a small (below one percentage) difference between the best model and the second-best model in the average overall score. These small difference in the overall performance metrics might not reflect any true difference in quality in the models. However, although the differences between the best performing models per model architecture was small, one consistent trait we found across the four model architectures was that models with expert-weighed cross entropy loss outperformed models with a weighed cross entropy loss. The expert-defined weight was implemented after the publication of Pazdernik et al. \cite{ pazdernik2020microstructural}, and consistently improved the model's overall performance. In addition, the Adam optimizer generally outperformed RMSopts optimizer. 

In regards to the microstructures, the predicted segmentation was similar to the expert-labels for pixel proportion, defect area, and defect density. For both, we found a higher average proportion of precipitates and voids across irradiated pellet images, smaller precipitate and void areas in two of the three irradiated pellet images, and a higher defect density for both precipitates and voids for all irradiated pellet images. For irradiated pellet images, we did, however, find a difference in whether the proportion of defects was significantly above or below 50\% for voids and impurities between the predicted and expert-labeled image. The lower performance on this microstucture quantification likely reflects the relatively low recall and precision of boundaries. Unfortunately, none of the additional adjustments (incorporating focal loss, topological loss, and meta-data) improved the average overall score or both the boundary recall and precision. This is a potential area for further study, perhaps with additional training images. 

Although there was a difference between the predicted and expert-label images for clustering and dispersions for impurities and voids at radiuses over 2.5 \(\mu m\) as measured by Ripley values, the predicted segmentation and the expert-labeled segmentation both found that voids clustered around precipitates, precipitates clustered around voids, and both voids and precipitates clustered around themselves between 0-2.5\(\mu m\). This shows that the predicted images has similar clustering and dispersion patterns as the expert-labeled image at smaller radiuses. The Ripley analysis of clustering of defects around other defects should provide a useful quantitative measure for what could otherwise be a subjective assessment of clustering. 

In regards to the microstructural quantification, there were some inconsistencies with observations noted in previous studies. Previous studies have noted a larger size of both voids and precipitates in irradiated images, which is somewhat inconsistent with our results \cite{jiang2020quantitative} \cite{jiang2022microstructural}  \cite{jiang2024microstructural}. The difference simply be due to our small sample size. Previous studies have noted that the size of precipitates is dependent on irradiation dose, temperature, and the microscope view direction \cite{jiang2024microstructural}, which also could explain any inconsistency between studies.

The decrease in precipitate proportion on grain boundaries in the irradiated pellet images is consistent with previous studies which found a defect-denuded zone at grain boundaries in irradiated pellets \cite{jiang2024microstructural}. Jiang et al. also found that some grain boundaries have a higher concentration of voids after irradiation \cite{jiang2024microstructural}. This finding is not necessarily inconsistent with our finding of a lower proportion of voids on grain boundaries compared with the proportion within the grains. The higher concentration of voids on the grain boundaries after irradiation could be due to an overall increase in the density of voids both on grain boundaries and within grains, instead of a migration of voids to the grain boundaries. Previous studies have in fact also noted a higher density of voids in irradiated pellets \cite{jiang2020quantitative}.

 In regards to the Ripley clustering results, considering that a previous paper had observed precipitates surrounded by voids in irradiated conditions \cite{jiang2022microstructural}, we were surprised not to see a more clear distinction between irradiated and unirradiated image pellets for clustering of precipitates and voids. Overall, it's difficult to draw any conclusions on microstructure evolution based on a small image number. However, training and utilizing a model like ours should make quantification and analysis of a much larger selection of images more feasible.  

There are many arguments against the use of machine learning techniques for scientific or other high precision tasks, many of which center around the ostensible lack of uncertainty quantification for such methods. This paper proposes a method to resolve that lack of uncertainty quantification for microstructural evaluation. Our novel contributions are the combination of a novel calibration method to enable confidence distribution output \cite{hagen2022dbcal} from a semantic segmentation network trained to segment microstructures from scanning electron micrographs \cite{ pazdernik2020microstructural}. We apply this novel pipeline to new data, and therefore provide both a quantitative and visual method to compare images of irradiated and unirradiated fuel pellets.
 
\section{Conclusion}

Our paper has shown the successful segmentation and quantification of microstrutures using DCNN for both irradiated and unirradiated pellet images. The high level of agreement between expert-labeled and predicted segmentation in most metrics for both pixel segmentation and microstructure analysis demonstrates that using DCNN is a feasible alternative to expert-labeling, although there are still metrics that could be improved with further training. In addition, we have provided a qualitative comparison of the microstructures of irradiated and unirradiated pellets. The methods described in this paper would be a helpful tool in obtaining quantitative information on pellet microstructures and predicting tritium-producing burnable absorber rod performance based on microstructural evolution.

\section{Funding}

This work was supported by the PNNL Tritium Technology Program, sponsored by the National Nuclear Security Administration (NNSA). The research was performed using resources available through Research Computing at Pacific Northwest National Laboratory (PNNL). PNNL is operated by Battelle for the U.S. Department of Energy under Contract DE-AC05-76RL01830.

\section{Data Availability}
The software is available at \url{https://github.com/pnnl/defect_detection}.The data is available at \url{https://data.pnnl.gov/group/nodes/dataset/34083}.

\section{Credit}
Marjolein Oostrom: Writing – original draft, Software, Visualization. Karl Pazdernik: Writing – review and editing, Software, Project Administration, Supervision, Conceptualization, Visualization. Alex Hagen: Writing – review and editing, Software, Conceptualization. Nicole LaHaye: Data curation

\section{Acknowledgements}
The authors would like to acknowledge Brad Johnson for providing the unirradiated LiAlO2 pellet SEM images, Edgar Buck for providing the irradiated LiAlO2 pellet SEM images, and Danny Edwards for discussions on the segmentation results.

\section{Conflict of Interest}
The authors declare that they have no conflict of interest.

 \clearpage 

\bibliographystyle{vancouver}
\bibliography{references}

\begin{thebibliography}{10}

\bibitem{senor2018science}
Senor D.
\newblock Science and technology in support of the tritium sustainment program.
\newblock PNNL report, PNNL-27216. 2018.

\bibitem{jiang2020quantitative}
Jiang W, Luscher WG, Wang T, Zhu Z, Shao L, Senor DJ.
\newblock A quantitative study of retention and release of deuterium and
  tritium during irradiation of $\gamma$-LiAlO2 pellets.
\newblock Journal of Nuclear Materials. 2020;542:152532.

\bibitem{jiang2022microstructural}
Jiang W, Kovarik L, Zhu Z, Varga T, Bowden ME, Matthews BE, et~al.
\newblock Microstructural evolution and precipitation in $\gamma$-LiAlO2 during
  ion irradiation.
\newblock Journal of Applied Physics. 2022;131(21).

\bibitem{jiang2024microstructural}
Jiang W, Kovarik L, Wirth MG, Hu Z, Shao L, Casella AM, et~al.
\newblock Microstructural and compositional evolutions in $\gamma$-LiAlO2
  pellets during ion irradiation at an elevated temperature.
\newblock Journal of Nuclear Materials. 2024;591:154925.

\bibitem{pazdernik2020microstructural}
Pazdernik K, LaHaye NL, Artman CM, Zhu Y.
\newblock Microstructural classification of unirradiated LiAlO2 pellets by deep
  learning methods.
\newblock Computational Materials Science. 2020;181:109728.

\bibitem{roberts2019deep}
Roberts G, Haile SY, Sainju R, Edwards DJ, Hutchinson B, Zhu Y.
\newblock Deep learning for semantic segmentation of defects in advanced STEM
  images of steels.
\newblock Scientific reports. 2019;9(1):12744.

\bibitem{jacobs2022performance}
Jacobs R, Shen M, Liu Y, Hao W, Li X, He R, et~al.
\newblock Performance and limitations of deep learning semantic segmentation of
  multiple defects in transmission electron micrographs.
\newblock Cell Reports Physical Science. 2022;3(5).

\bibitem{li2024machine}
Li X, Mai Y, Meng H, Bi H, Ng CH, Teo SH, et~al.
\newblock Machine learning quantification of grain boundary defects for high
  efficiency perovskite solar cells.
\newblock Advanced Composites and Hybrid Materials. 2024;7(6):1-8.

\bibitem{kirillov2023segment}
Kirillov A, Mintun E, Ravi N, Mao H, Rolland C, Gustafson L, et~al.
\newblock Segment anything.
\newblock In: Proceedings of the IEEE/CVF International Conference on Computer
  Vision; 2023. p. 4015-26.

\bibitem{zou2024segment}
Zou X, Yang J, Zhang H, Li F, Li L, Wang J, et~al.
\newblock Segment everything everywhere all at once.
\newblock Advances in Neural Information Processing Systems. 2024;36.

\bibitem{he2016deep}
He K, Zhang X, Ren S, Sun J.
\newblock Deep residual learning for image recognition.
\newblock In: Proceedings of the IEEE conference on computer vision and pattern
  recognition; 2016. p. 770-8.

\bibitem{deng2009imagenet}
Deng J, Dong W, Socher R, Li LJ, Li K, Fei-Fei L.
\newblock Imagenet: A large-scale hierarchical image database.
\newblock In: 2009 IEEE conference on computer vision and pattern recognition.
  Ieee; 2009. p. 248-55.

\bibitem{dosovitskiy2020image}
Dosovitskiy A, Beyer L, Kolesnikov A, Weissenborn D, Zhai X, Unterthiner T,
  et~al.
\newblock An image is worth 16x16 words: Transformers for image recognition at
  scale.
\newblock arXiv preprint arXiv:201011929. 2020.

\bibitem{liu2021efficient}
Liu Y, Sangineto E, Bi W, Sebe N, Lepri B, Nadai M.
\newblock Efficient training of visual transformers with small datasets.
\newblock Advances in Neural Information Processing Systems. 2021;34:23818-30.

\bibitem{huang2024segment}
Huang Y, Yang X, Liu L, Zhou H, Chang A, Zhou X, et~al.
\newblock Segment anything model for medical images?
\newblock Medical Image Analysis. 2024;92:103061.

\bibitem{badrinarayanan2017segnet}
Badrinarayanan V, Kendall A, Cipolla R.
\newblock Segnet: A deep convolutional encoder-decoder architecture for image
  segmentation.
\newblock IEEE transactions on pattern analysis and machine intelligence.
  2017;39(12):2481-95.

\bibitem{ronneberger2015u}
Ronneberger O, Fischer P, Brox T.
\newblock U-net: Convolutional networks for biomedical image segmentation.
\newblock In: Medical image computing and computer-assisted
  intervention--MICCAI 2015: 18th international conference, Munich, Germany,
  October 5-9, 2015, proceedings, part III 18. Springer; 2015. p. 234-41.

\bibitem{say4n}
say4n. pytorch-segnet: Implementation of SegNet architecture in Pytorch.
  GitHub; 2020.
\newblock \url{https://github.com/say4n/pytorch-segnet/tree/master}.

\bibitem{kendall2015bayesian}
Kendall A, Badrinarayanan V, Cipolla R.
\newblock Bayesian segnet: Model uncertainty in deep convolutional
  encoder-decoder architectures for scene understanding.
\newblock arXiv preprint arXiv:151102680. 2015.

\bibitem{srivastava2014dropout}
Srivastava N, Hinton G, Krizhevsky A, Sutskever I, Salakhutdinov R.
\newblock Dropout: a simple way to prevent neural networks from overfitting.
\newblock The journal of machine learning research. 2014;15(1):1929-58.

\bibitem{Ganaye}
Ganaye PA. pytorch-unet-segnet. GitHub; 2022.
\newblock \url{https://github.com/trypag/pytorch-unet-segnet/tree/master}.

\bibitem{zhang2022resnest}
Zhang H, Wu C, Zhang Z, Zhu Y, Lin H, Zhang Z, et~al.
\newblock Resnest: Split-attention networks.
\newblock In: Proceedings of the IEEE/CVF Conference on Computer Vision and
  Pattern Recognition; 2022. p. 2736-46.

\bibitem{trypag}
trypag. pytorch-unet-segnet. GitHub; 2019.
\newblock \url{https://github.com/trypag/pytorch-unet-segnet/tree/master}.

\bibitem{hagen2022dbcal}
Hagen A, Pazdernik K, LaHaye N, Oostrom M.
\newblock DBCal: Density based calibration of classifier predictions for
  uncertainty quantification.
\newblock arXiv preprint arXiv:220400150. 2022.

\bibitem{borges2019physics}
Borges P, Sudre C, Varsavsky T, Thomas D, Drobnjak I, Ourselin S, et~al.
\newblock Physics-informed brain MRI segmentation.
\newblock In: International Workshop on Simulation and Synthesis in Medical
  Imaging. Springer; 2019. p. 100-9.

\bibitem{pazdernik_microstructural_2020}
Pazdernik K, LaHaye NL, Artman CM, Zhu Y.
\newblock Microstructural classification of unirradiated {LiAlO2} pellets by
  deep learning methods.
\newblock Computational Materials Science. 2020 Aug;181:109728.
\newblock Available from:
  \url{http://www.sciencedirect.com/science/article/pii/S0927025620302196}.

\bibitem{lin2017focal}
Lin TY, Goyal P, Girshick R, He K, Doll{\'a}r P.
\newblock Focal loss for dense object detection.
\newblock In: Proceedings of the IEEE international conference on computer
  vision; 2017. p. 2980-8.

\bibitem{AdeelH}
AdeelH. Multi-class Focal Loss. GitHub; 2022.
\newblock \url{https://github.com/AdeelH/pytorch-multi-class-focal-loss}.

\bibitem{hu2019topology}
Hu X, Li F, Samaras D, Chen C.
\newblock Topology-preserving deep image segmentation.
\newblock Advances in neural information processing systems. 2019;32.

\bibitem{scikit-learn}
Pedregosa F, Varoquaux G, Gramfort A, Michel V, Thirion B, Grisel O, et~al.
\newblock Scikit-learn: Machine Learning in {P}ython.
\newblock Journal of Machine Learning Research. 2011;12:2825-30.

\bibitem{nelson2017assessing}
Nelson KP.
\newblock Assessing probabilistic inference by comparing the generalized mean
  of the model and source probabilities.
\newblock Entropy. 2017;19(6):286.

\bibitem{opencv_library}
Bradski G.
\newblock {The OpenCV Library}.
\newblock Dr Dobb's Journal of Software Tools. 2000.

\bibitem{ripley1977modelling}
Ripley BD.
\newblock Modelling spatial patterns.
\newblock Journal of the Royal Statistical Society: Series B (Methodological).
  1977;39(2):172-92.

\bibitem{baddeley2005spatstat}
Baddeley A, Turner R.
\newblock Spatstat: an R package for analyzing spatial point patterns.
\newblock Journal of statistical software. 2005;12:1-42.

\bibitem{2022ApJ...935..167A}
{Astropy Collaboration}, {Price-Whelan} AM, {Lim} PL, {Earl} N, {Starkman} N,
  {Bradley} L, et~al.
\newblock {The Astropy Project: Sustaining and Growing a Community-oriented
  Open-source Project and the Latest Major Release (v5.0) of the Core Package}.
\newblock \apj. 2022 Aug;935(2):167.

\bibitem{marcon2009generalizing}
Marcon E, Puech F.
\newblock Generalizing Ripley's K function to inhomogeneous populations. 2009.

\bibitem{getis1984interaction}
Getis A.
\newblock Interaction modeling using second-order analysis.
\newblock Environment and Planning A. 1984;16(2):173-83.

\bibitem{dixon2012ripley}
Dixon PM, El-Shaarawi AH, Piegorsch WW.
\newblock Ripley’s K function.
\newblock Encyclopedia of Environmetrics. 2012;3.

\end{thebibliography}
\newpage 
\setcounter{table}{0}
\renewcommand{\thetable}{A\arabic{table}}
\begingroup
\squeezetable
\begin{table}

\caption {\label{tab:table_performance_metrics_unirradiated_weighted}  Performance Metrics for Unirradiated Pellets with Weighted Metrics: Precision, recall, and \(F_{D}\) for pixel metrics, and Box IoU precision (\(Box_{p}\)), Box IoU recall (\(Box_{r}\)), average of Box recall and Precision (\(Box_{a}\)), and the average of the \(F_{D}\), IoU, and Box A scores for irradiated pellet images. Metrics are for the Bayesian SegNeSt (BayesSegNeSt), Bayesian SegNet (Bayes), SegNet, and Small Bayesian SegNet (SmallBayes) models and  using either a EWCE or WCE loss and Adam and RMS optimizer. The models used for the additional adjustments are a SegNet model with a Adam optimizer and Focal (FL) or Topological (Topo) loss.}

\begin{ruledtabular}
\begin{tabular}{lll|llll|llll}
Models & Opt. & Loss & Precision & Recall &  \(F_{D}\) & IoU & \(Box_{p}\) & \(Box_{r}\) & \(Box_{a}\) & Avg \\    
 \hline
 \multirow{4}{*}{BayesSegNeSt}   &  \multirow{2}{*}{Adam}   & EWCE  & 96.2\% & 95.7\% & \textbf{96.8\%} & 92.7\% & 54.2\% & 61.9\% & 58.1\% & 82.5\%  \\ 
 \cline{3-11} 
 &     & WCE  & 95.6\% & 93.8\% & 96.4\% & 90.1\% & 43.8\% & 53.7\% & 48.8\% & 78.5\%   \\  
  \cline{3-11} 
 &   \multirow{2}{*}{RMS}   & EWCE  & 96.2\% & 95.7\% & 96.7\% & 92.7\% & \textbf{56.7\%} & 62\% & 59.3\% & 82.9\% \\ 
 \cline{3-11}   
  &     & WCE  & 95.4\% & 94.1\% & 96.2\% & 90.4\% & 47.9\% & 57\% & 52.4\% & 79.7\%  \\  
\hline
 \multirow{4}{*}{Bayes SegNet}   &  \multirow{2}{*}{Adam}   & EWCE  & 96.1\% & 95.6\% & 96.5\% & 92.4\% & 54.6\% & 63\% & 58.8\% & 82.6\%  \\ 
 \cline{3-11} 
 &     & WCE  & 95.1\% & 92.9\% & 95.8\% & 89\% & 43\% & 53.3\% & 48.2\% & 77.7\% \\  
  \cline{3-11} 
 &   \multirow{2}{*}{RMS}   & EWCE  & 95.8\% & 95.3\% & 96.2\% & 92\% & 52.5\% & 60.8\% & 56.6\% & 81.6\% \\ 
 \cline{3-11}   
  &     & WCE  & 95.2\% & 92.6\% & 95.8\% & 88.7\% & 42.6\% & 54.7\% & 48.6\% & 77.7\%   \\  
  \hline
 \multirow{4}{*}{SegNet}   &  \multirow{2}{*}{Adam}   & EWCE  & \textbf{96.3\%} & \textbf{96.2\%} & 96.5\% & \textbf{93.2\%} & 55.9\% & 64.6\% & 60.3\% & \textbf{83.3\%}   \\ 
 \cline{3-11} 
 &     & WCE  & 96.3\% & 96\% & 96.5\% & 93\% & 56.3\% & 64.3\% & 60.3\% & 83.3\% \\  
  \cline{3-11} 
 &   \multirow{2}{*}{RMS}   & EWCE  & 96.1\% & 96.1\% & 96.2\% & 93.1\% & 54.2\% & \textbf{64.7\%} & 59.5\% & 82.9\%  \\ 
 \cline{3-11}   
  &     & WCE  & 96.1\% & 95.9\% & 96.3\% & 92.8\% & 54.2\% & 63.6\% & 58.9\% & 82.7\%     \\  
  \hline
 \multirow{4}{*}{Small Bayes}   &  \multirow{2}{*}{Adam}   & EWCE  & 96.1\% & 95.6\% & 96.6\% & 92.5\% & 56.6\% & 62.5\% & \textbf{59.6\%} & 82.9\%  \\ 
 \cline{3-11} 
 &     & WCE  & 95.5\% & 93.7\% & 96.2\% & 90\% & 41.3\% & 54.2\% & 47.7\% & 78\%  \\  
  \cline{3-11} 
 &   \multirow{2}{*}{RMS}   & EWCE  & 96\% & 95.6\% & 96.6\% & 92.4\% & 54\% & 62.4\% & 58.2\% & 82.4\%  \\ 
 \cline{3-11}   
  &     & WCE  & 95.3\% & 93.7\% & 96.1\% & 90\% & 41.6\% & 54.6\% & 48.1\% & 78\% \\  
 \hline
 \multirow{2}{*}{SmallBayes} & \multirow{2}{*}{Adam} & Focal Loss  & 96.1\% & 96\% & 96.2\% & 93\% & 54.3\% & 62.7\% & 58.5\% & 82.6\%  \\ 

 & & Topology  & 96.2\% & 96.1\% & 96.3\% & 93.1\% & 54.2\% & 63.4\% & 58.8\% & 82.8\%  \\ 

\end{tabular}
\end{ruledtabular}
\end{table}
\endgroup

\begingroup
\squeezetable
\begin{table}[!ht]
    
    \caption {\label{tab:table_performance_metrics_irradiated_weighted}  Performance Metrics for Irradiated Pellets with Weighted Metrics: Precision, recall, and \(F_{D}\) for pixel metrics, and Box IoU precision (\(Box_{p}\)), Box IoU recall (\(Box_{r}\)), average of Box recall and Precision (\(Box_{a}\)), and the average of the \(F_{D}\), IoU, and Box A scores for irradiated pellet images. Metrics are for the Bayesian SegNeSt (BayesSegNeSt), Bayesian SegNet (Bayes), SegNet, and Small Bayesian SegNet (SmallBayes) models and  using either a Expert Weighted Cross Entropy (EWCE) or Weighted Cross-Entropy (WCE) loss and Adam and RMS optimizer for the initial models. The models used for the additional adjustments are a Small Bayesian SegNet (SmallBayes) model with a Adam optimizer and Focal (FL) or Topological (Topo) loss or a SegNet model with metadata and a Adam optimizer and EWCE loss. The best score for each metric is in bold.}

\begin{ruledtabular}
\begin{tabular}{lll|llll|llll}
Models & Opt. & Loss & Precision & Recall &  \(F_{D}\) & IoU & \(Box_{p}\) & \(Box_{r}\) & \(Box_{a}\) & Avg \\    
 \hline
 \multirow{4}{*}{BayesSegNeSt}   &  \multirow{2}{*}{Adam}   & EWCE  & \textbf{94.2\%} & 94\% & 95.3\% & 89\% & 50.6\% & 51.2\% & 50.9\% & 78.4\% \\ 
 \cline{3-11} 
 &     & WCE  & 93.4\% & 91.1\% & 94.2\% & 86.2\% & 50.3\% & 49.2\% & 49.8\% & 76.7\%    \\  
  \cline{3-11} 
 &   \multirow{2}{*}{RMS}   & EWCE  & 94\% & 93\% & \textbf{95.3\%} & 87.8\% & 46.8\% & 39.6\% & 42.5\% & 75.2\%  \\ 
 \cline{3-11}   
  &     & WCE  & 93\% & 90.8\% & 93.6\% & 85.7\% & 49.8\% & 44.6\% & 47.2\% & 75.5\% \\  
\hline
 \multirow{4}{*}{Bayes SegNet}   &  \multirow{2}{*}{Adam}   & EWCE  & 93.7\% & 92.9\% & 94.2\% & 88.4\% & 61.2\% & 52.1\% & 56.7\% & 79.7\%  \\ 
 \cline{3-11} 
 &     & WCE  & 91.3\% & 87.4\% & 92.3\% & 81.1\% & 38.3\% & 46.3\% & 42.3\% & 71.9\% \\  
  \cline{3-11} 
 &   \multirow{2}{*}{RMS}   & EWCE  & 93.4\% & 92.4\% & 95.1\% & 86.6\% & 45.4\% & 39.2\% & 41.7\% & 74.5\%   \\ 
 \cline{3-11}   
  &     & WCE  & 92.8\% & 76\% & 89.1\% & 71\% & 38.9\% & 47.6\% & 43.3\% & 67.8\%   \\  
  \hline
 \multirow{4}{*}{SegNet}   &  \multirow{2}{*}{Adam}   & EWCE  & 93.7\% & \textbf{94.1\%} & 93.4\% & 89.5\% & 55.8\% & 54.3\% & 55\% & 79.3\% \\ 
 \cline{3-11} 
 &     & WCE  & 93.3\% & 91.8\% & 93.8\% & 87\% & 49\% & 48.9\% & 48.9\% & 76.6\% \\  
  \cline{3-11} 
 &   \multirow{2}{*}{RMS}   & EWCE  & 93.7\% & 93.7\% & 93.6\% & 89.2\% & 53.7\% & 50.9\% & 52.3\% & 78.4\% \\ 
 \cline{3-11}   
  &     & WCE  & 93.6\% & 93.8\% & 93.3\% & 89.3\% & 50.4\% & 52\% & 51.2\% & 77.9\%    \\  
  \hline

 \multirow{4}{*}{Small Bayes}   &  \multirow{2}{*}{Adam}   & EWCE  & 93.9\% & 92.8\% & 94.3\% & 88.3\% & \textbf{63.3\%} & 51.8\% & \textbf{57.5\%} & \textbf{80\%} \\ 
 \cline{3-11} 
 &     & WCE  & 93.7\% & 89\% & 93.7\% & 84.4\% & 45.9\% & 49.4\% & 47.7\% & 75.3\% \\  
  \cline{3-11} 
 &   \multirow{2}{*}{RMS}   & EWCE  & 93.7\% & 92.6\% & 94.3\% & 88\% & 50.6\% & 52.1\% & 51.4\% & 77.9\% \\ 
 \cline{3-11}   
  &     & WCE  & 93.5\% & 87.7\% & 92.8\% & 83.1\% & 43.5\% & 48.3\% & 45.9\% & 73.9\%   \\  
 \hline
 \multirow{2}{*}{SmallBayes} & \multirow{2}{*}{Adam} & Focal Loss  & 92.4\% & 92.4\% & 93.8\% & 86.5\% & 47.1\% & 39.7\% & 42.7\% & 74.3\%  \\ 

 & & Topology  & 93.8\% & 92.7\% & 94.4\% & 88.1\% & 53.8\% & 50.8\% & 52.3\% & 78.3\%  \\ 

 SegNet Meta &  Adam  & EWCE  & 93.9\% & 94\% & 93.9\% & \textbf{89.5\%} & 57.2\% & \textbf{54.4\%} & 55.8\% & 79.7\% \\ 
\end{tabular}
\end{ruledtabular}
\end{table}
\endgroup

\begingroup
\squeezetable
\begin{table}

 \caption {\label{tab:additional_meta_data} 
 Additional meta-data information per irradiated image.}
    
    \begin{ruledtabular}
    \begin{tabular}{|l|l|l|l|l|}
    \hline
        \textbf{Additional meta-data} & \textbf{Units}& \textbf{Image 3} & \textbf{Image 4}  & \textbf{Image 5}  \\ \hline
        Time & Hours &  11 & 10 & 9 \\ \hline
        Date & Days since Jan 1, 2017 &  311 & 445 & 138 \\ \hline
        Image Type (SEM) & boolean & 1 & 1 & 1 \\ \hline
        Accelerating Voltage & kV & 10 & 5 & 15 \\ \hline
        Beam spot & um &  5.5 & 6.5 & 6 \\ \hline
        Beam Type (ebeam) & boolean &  1 & 1 & 1 \\ \hline
        Scan Type (escan) & boolean &   1 & 1 & 1 \\ \hline
        Electron Accelerating Voltage & kV &  10 & 5 & 15 \\ \hline
        Horizontal Field Width & um & 29.8 & 29.8 & 19.9 \\ \hline
        Vertical Field Width & um & 25.8 & 25.8 & 17.2 \\ \hline
        Beam Current & pA & 158 & 178 & 142  \\ \hline
        Dwell Time & micro-seconds &  5 & 10 & 3 \\ \hline
        Signal Type (Backscattered-Electron) & boolean &  1 & 1 & 1 \\ \hline
        Location of Data Bar & pixels & 119 & 119 & 119 \\ \hline
        Magnification & x & 10000 & 10000 & 15000 \\ \hline

\end{tabular}
\end{ruledtabular}
\end{table}

\endgroup

\begingroup
\squeezetable
\begin{table}

 \caption {\label{tab:prediction_proportion} Prediction Proportion Table: The number of images (n), the mean proportion of defect (Prop), standard deviation (STD), and standard error (SE) per defect for irradiated and unirradiated pellet images (Condition) and expert-labeled (T) and predicted images (P).}

    \begin{ruledtabular}
    \begin{tabular}{|l|l|l|l|l|l|l|}
    \hline
        \textbf{Condition} & \textbf{Defect} & \textbf{Type} & \textbf{n} & \textbf{Prop} & \textbf{STD} & \textbf{SE} \\ \hline
                Irradiated & Boundary & P & 3 & 0.03 & 0.004 & 0.005 \\ \hline
        Irradiated & Boundary & T & 3 & 0.018 & 0.003 & 0.003 \\ \hline
        Irradiated & Grain & P & 3 & 0.787 & 0.097 & 0.11 \\ \hline
        Irradiated & Grain & T & 3 & 0.811 & 0.102 & 0.116 \\ \hline
        Irradiated & Impurity & P & 3 & 0.001 & 0.001 & 0.001 \\ \hline
        Irradiated & Impurity & T & 3 & 0.003 & 0.001 & 0.001 \\ \hline
        Irradiated & Precipitate & P & 3 & 0.09 & 0.063 & 0.071 \\ \hline
        Irradiated & Precipitate & T & 3 & 0.078 & 0.059 & 0.067 \\ \hline
        Irradiated & Void & P & 3 & 0.093 & 0.065 & 0.073 \\ \hline
        Irradiated & Void & T & 3 & 0.091 & 0.069 & 0.079 \\ \hline
        Unirradiated & Boundary & P & 2 & 0.049 & 0.004 & 0.005 \\ \hline
        Unirradiated & Boundary & T & 2 & 0.046 & 0.004 & 0.005 \\ \hline
        Unirradiated & Grain & P & 2 & 0.875 & 0.01 & 0.014 \\ \hline
        Unirradiated & Grain & T & 2 & 0.88 & 0.011 & 0.016 \\ \hline
        Unirradiated & Impurity & P & 2 & 0.006 & 0 & 0 \\ \hline
        Unirradiated & Impurity & T & 2 & 0.005 & 0 & 0 \\ \hline
        Unirradiated & Precipitate & P & 2 & 0.031 & 0.001 & 0.002 \\ \hline
        Unirradiated & Precipitate & T & 2 & 0.031 & 0.002 & 0.003 \\ \hline
        Unirradiated & Void & P & 2 & 0.039 & 0.007 & 0.01 \\ \hline
        Unirradiated & Void & T & 2 & 0.038 & 0.01 & 0.013 \\ \hline
    \end{tabular}
\end{ruledtabular}
\end{table}
\endgroup

\begingroup
\squeezetable
\begin{table}

    \caption{\label{tab:table_average_area_density} Average Area and Density: The number of defects (n), mean area (Area), standard deviation of the area (STD), standard error (SE), and density of defects (Density) per the expert-labeled (Truth) for irradiated and unirradiated pellet images (Condition) and expert-labeled (T) and predicted images (P).}

        \begin{ruledtabular}
    \begin{tabular}{|l|l|l|l|l|l|l|l|l|}
    \hline
        \textbf{Condition} & \textbf{Image} & \textbf{Defect} & \textbf{Pred} & \textbf{n} & \textbf{Area} & \textbf{STD} & \textbf{SE} & \textbf{Density} \\ \hline
        Irradiated & 3 & Impurity & T & 15 & 0.023 & 0.017 & 0.009 & 0.123 \\ \hline
        Irradiated & 3 & Precipitate & T & 387 & 0.037 & 0.07 & 0.007 & 3.176 \\ \hline
        Irradiated & 3 & Precipitate & P & 424 & 0.039 & 0.072 & 0.007 & 3.48 \\ \hline
        Irradiated & 3 & Void & T & 258 & 0.032 & 0.051 & 0.006 & 2.117 \\ \hline
        Irradiated & 3 & Void & P & 270 & 0.031 & 0.052 & 0.006 & 2.216 \\ \hline
        Irradiated & 4 & Impurity & T & 12 & 0.041 & 0.027 & 0.016 & 0.069 \\ \hline
        Irradiated & 4 & Impurity & P & 6 & 0.057 & 0.027 & 0.022 & 0.035 \\ \hline
        Irradiated & 4 & Precipitate & T & 133 & 0.024 & 0.032 & 0.005 & 0.769 \\ \hline
        Irradiated & 4 & Precipitate & P & 163 & 0.028 & 0.043 & 0.007 & 0.942 \\ \hline
        Irradiated & 4 & Void & T & 148 & 0.027 & 0.053 & 0.009 & 0.855 \\ \hline
        Irradiated & 4 & Void & P & 180 & 0.029 & 0.056 & 0.008 & 1.04 \\ \hline
        Irradiated & 5 & Impurity & T & 3 & 0.043 & 0.041 & 0.047 & 0.035 \\ \hline
        Irradiated & 5 & Impurity & P & 2 & 0.052 & 0.056 & 0.078 & 0.023 \\ \hline
        Irradiated & 5 & Precipitate & T & 53 & 0.11 & 0.192 & 0.052 & 0.621 \\ \hline
        Irradiated & 5 & Precipitate & P & 51 & 0.132 & 0.201 & 0.055 & 0.597 \\ \hline
        Irradiated & 5 & Void & T & 110 & 0.124 & 0.346 & 0.065 & 1.288 \\ \hline
        Irradiated & 5 & Void & P & 119 & 0.113 & 0.292 & 0.052 & 1.393 \\ \hline
        Unirradiated & 1 & Impurity & T & 220 & 0.011 & 0.011 & 0.001 & 0.328 \\ \hline
        Unirradiated & 1 & Impurity & P & 250 & 0.01 & 0.01 & 0.001 & 0.373 \\ \hline
        Unirradiated & 1 & Precipitate & T & 110 & 0.181 & 0.248 & 0.046 & 0.164 \\ \hline
        Unirradiated & 1 & Precipitate & P & 105 & 0.185 & 0.261 & 0.05 & 0.157 \\ \hline
        Unirradiated & 1 & Void & T & 109 & 0.172 & 0.341 & 0.064 & 0.162 \\ \hline
        Unirradiated & 1 & Void & P & 201 & 0.098 & 0.215 & 0.03 & 0.3 \\ \hline
        Unirradiated & 2 & Impurity & T & 239 & 0.01 & 0.011 & 0.001 & 0.356 \\ \hline
        Unirradiated & 2 & Impurity & P & 243 & 0.011 & 0.01 & 0.001 & 0.362 \\ \hline
        Unirradiated & 2 & Precipitate & T & 126 & 0.139 & 0.179 & 0.031 & 0.188 \\ \hline
        Unirradiated & 2 & Precipitate & P & 132 & 0.138 & 0.229 & 0.039 & 0.197 \\ \hline
        Unirradiated & 2 & Void & T & 198 & 0.139 & 0.343 & 0.048 & 0.295 \\ \hline
        Unirradiated & 2 & Void & P & 255 & 0.109 & 0.27 & 0.033 & 0.38 \\ \hline
    \end{tabular}
\end{ruledtabular}
\end{table}
\endgroup

\begingroup
\squeezetable
\begin{table}

    \caption {\label{tab:prop_grain_boundary_table} Percentages and number (n) of defects on boundary (onboundary) or in grain (ingrain), total number (N), standard error (SE), and significance (Sig) for irradiated and unirradiated pellet images (Condition) and expert-labeled (T) and predicted images (P).} 
        
        \begin{ruledtabular}
    \begin{tabular}{|l|l|l|l|l|l|l|l|l|l|}
    \hline
        \textbf{Defect} & \textbf{Pred} & \textbf{Condition} & \textbf{Type} & \textbf{n} & \textbf{N} & \textbf{Proportion} & \textbf{SE} & \textbf{P-value} & Sig \\ \hline
        Impurity & T & Irradiated & ingrain & 18 & 30 & 0.6 & 0.175 & 0.451 & FALSE \\ \hline
        Impurity & T & Irradiated & onboundary & 12 & 30 & 0.4 & 0.175 & 0.451 & FALSE \\ \hline
        Impurity & T & Unirradiated & ingrain & 314 & 459 & 0.684 & 0.043 & 0.451 & FALSE \\ \hline
        Impurity & T & Unirradiated & onboundary & 145 & 459 & 0.316 & 0.043 & 0.451 & FALSE \\ \hline
        Impurity & P & Irradiated & ingrain & 2 & 8 & 0.25 & 0.3 & 0.037 & TRUE \\ \hline
        Impurity & P & Irradiated & onboundary & 6 & 8 & 0.75 & 0.3 & 0.037 & TRUE \\ \hline
        Impurity & P & Unirradiated & ingrain & 328 & 493 & 0.665 & 0.042 & 0.037 & TRUE \\ \hline
        Impurity & P & Unirradiated & onboundary & 165 & 493 & 0.335 & 0.042 & 0.037 & TRUE \\ \hline
        Precipitate & T & Irradiated & ingrain & 418 & 573 & 0.729 & 0.036 & 0 & TRUE \\ \hline
        Precipitate & T & Irradiated & onboundary & 155 & 573 & 0.271 & 0.036 & 0 & TRUE \\ \hline
        Precipitate & T & Unirradiated & ingrain & 28 & 236 & 0.119 & 0.041 & 0 & TRUE \\ \hline
        Precipitate & T & Unirradiated & onboundary & 208 & 236 & 0.881 & 0.041 & 0 & TRUE \\ \hline
        Precipitate & P & Irradiated & ingrain & 348 & 638 & 0.545 & 0.039 & 0 & TRUE \\ \hline
        Precipitate & P & Irradiated & onboundary & 290 & 638 & 0.455 & 0.039 & 0 & TRUE \\ \hline
        Precipitate & P & Unirradiated & ingrain & 17 & 237 & 0.072 & 0.033 & 0 & TRUE \\ \hline
        Precipitate & P & Unirradiated & onboundary & 220 & 237 & 0.928 & 0.033 & 0 & TRUE \\ \hline
        Void & T & Irradiated & ingrain & 335 & 516 & 0.649 & 0.041 & 0 & TRUE \\ \hline
        Void & T & Irradiated & onboundary & 181 & 516 & 0.351 & 0.041 & 0 & TRUE \\ \hline
        Void & T & Unirradiated & ingrain & 93 & 307 & 0.303 & 0.051 & 0 & TRUE \\ \hline
        Void & T & Unirradiated & onboundary & 214 & 307 & 0.697 & 0.051 & 0 & TRUE \\ \hline
        Void & P & Irradiated & ingrain & 234 & 569 & 0.411 & 0.04 & 0 & TRUE \\ \hline
        Void & P & Irradiated & onboundary & 335 & 569 & 0.589 & 0.04 & 0 & TRUE \\ \hline
        Void & P & Unirradiated & ingrain & 84 & 456 & 0.184 & 0.036 & 0 & TRUE \\ \hline
        Void & P & Unirradiated & onboundary & 372 & 456 & 0.816 & 0.036 & 0 & TRUE \\ \hline
    \end{tabular}
\end{ruledtabular}
\end{table}
\endgroup

\setcounter{figure}{0}
\renewcommand{\thefigure}{A\arabic{figure}}
\begin{figure}[htbp]
    
    \includegraphics[width=\textwidth]{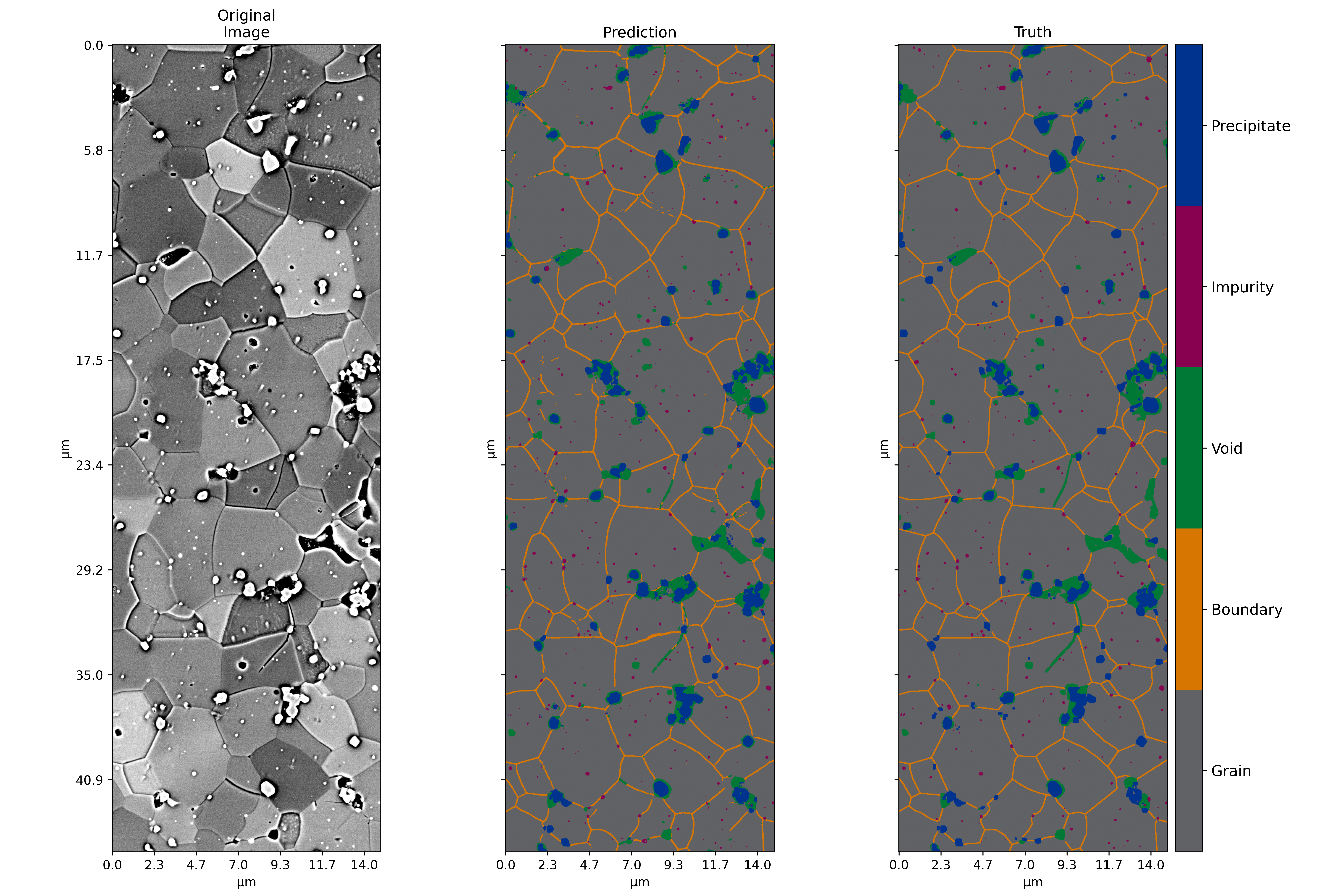}
      \caption{A\label{fig:S1_unirradiated_image_1} Predictions: From left to right, 1) the original test section of ``Image 1'' which was unirradiated, 2) the predicted image, and 3) the expert-labeled image. The predicted images were created with these parameters: SegNet model architecture, EWCE loss, and Adam optimizer.}
    
\end{figure}

\begin{figure}[htbp]
    
    \includegraphics[width=\textwidth]{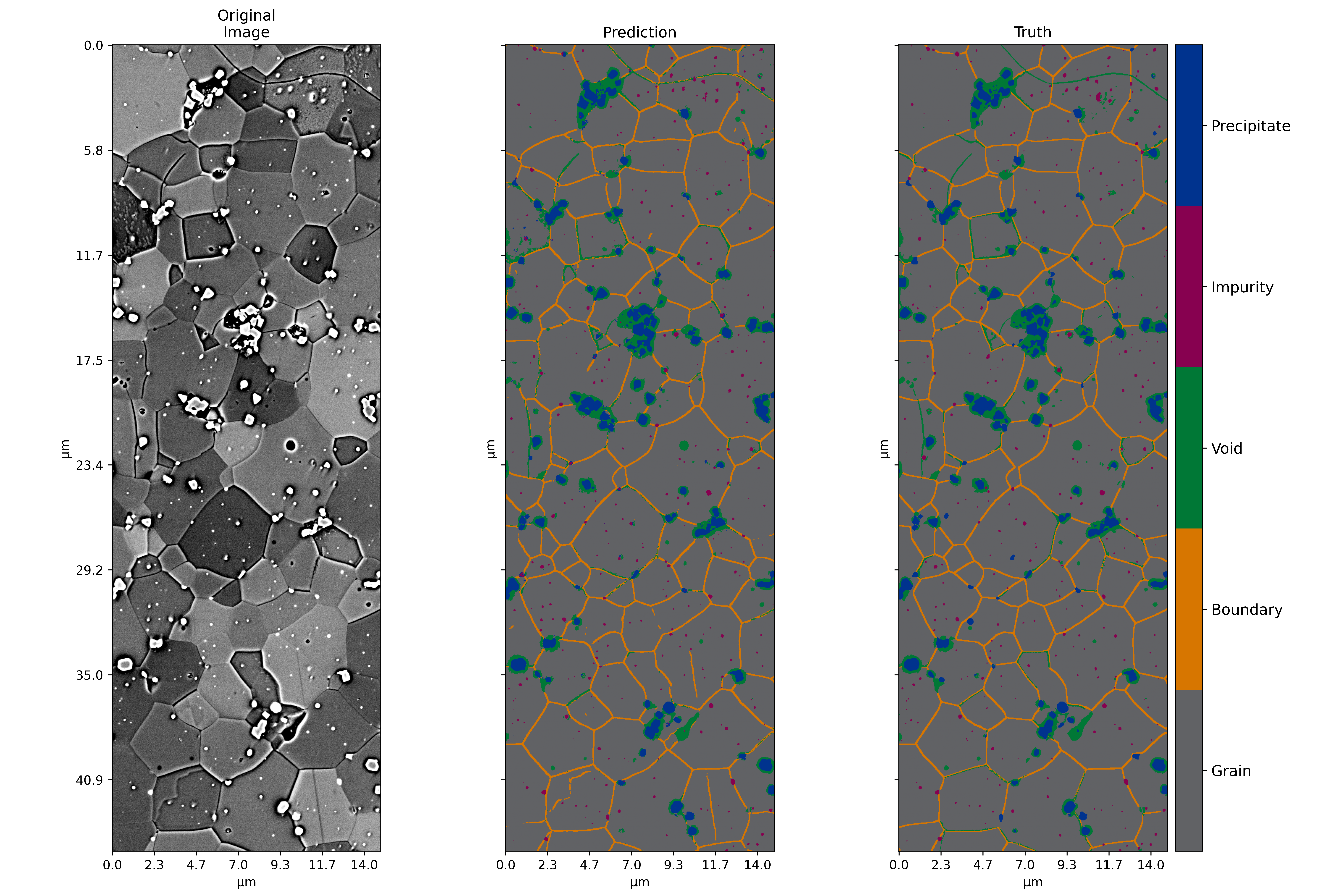}
      \caption{\label{fig:S2_unirradiated_image_2} Predictions: From left to right, 1) the original test section of ``Image 2'' which was unirradiated, 2) the predicted image, and 3) the expert-labeled image. The predicted images were created with these parameters: SegNet model architecture, EWCE loss, and Adam optimizer.}
    
\end{figure}

\begin{figure}[htbp]
    
    \includegraphics[width=\textwidth]{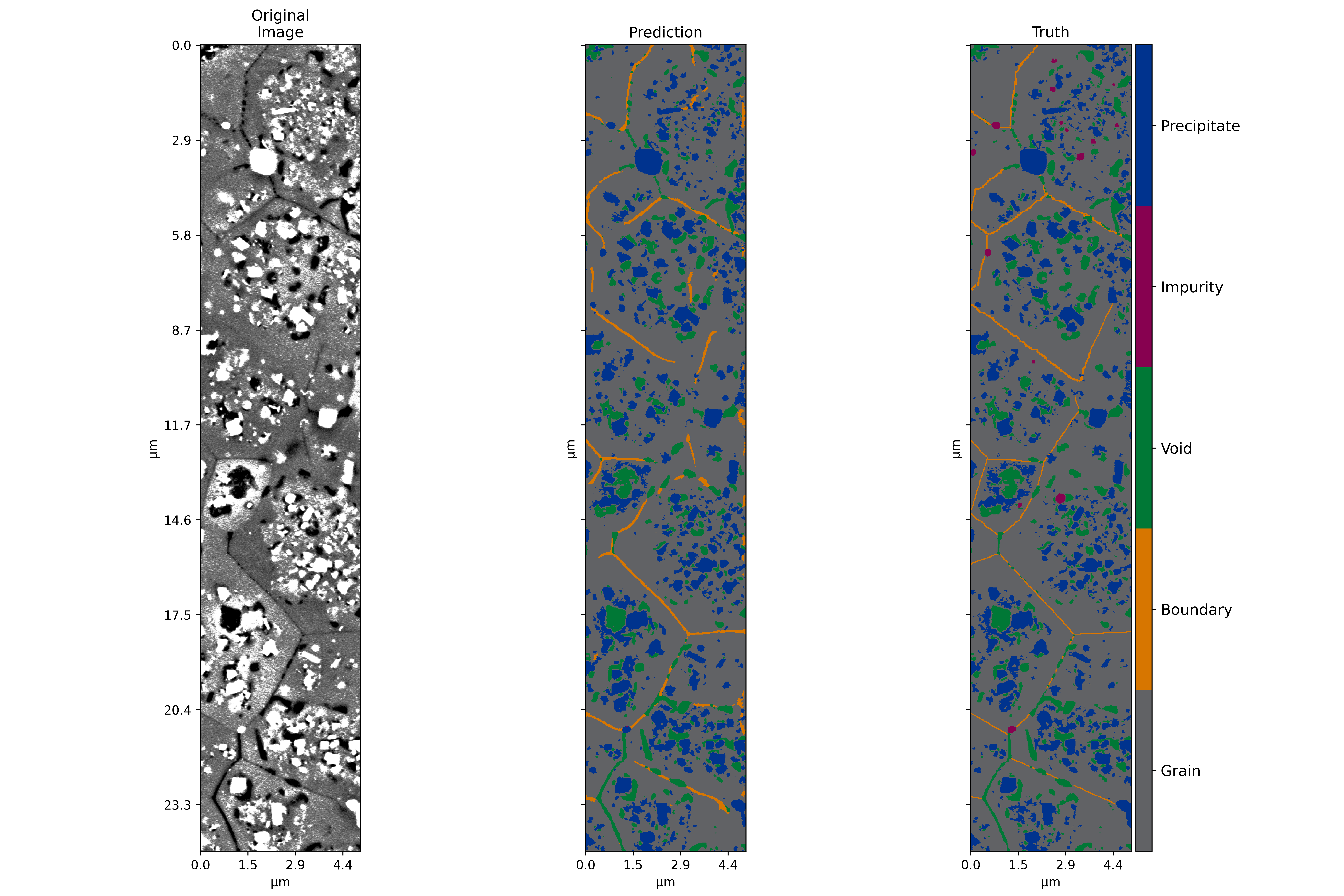}
      \caption{\label{fig:S3_unirradiated_image_3} Predictions: From left to right, 1) the original test section of ``Image 3'' which was irradiated, 2) the predicted image, and 3) the expert-labeled image. The predicted images were created with these parameters: Small Bayesian SegNet model architecture, EWCE loss, and Adam optimizer.}
    
\end{figure}

\begin{figure}[htbp]
    
    \includegraphics[width=\textwidth]{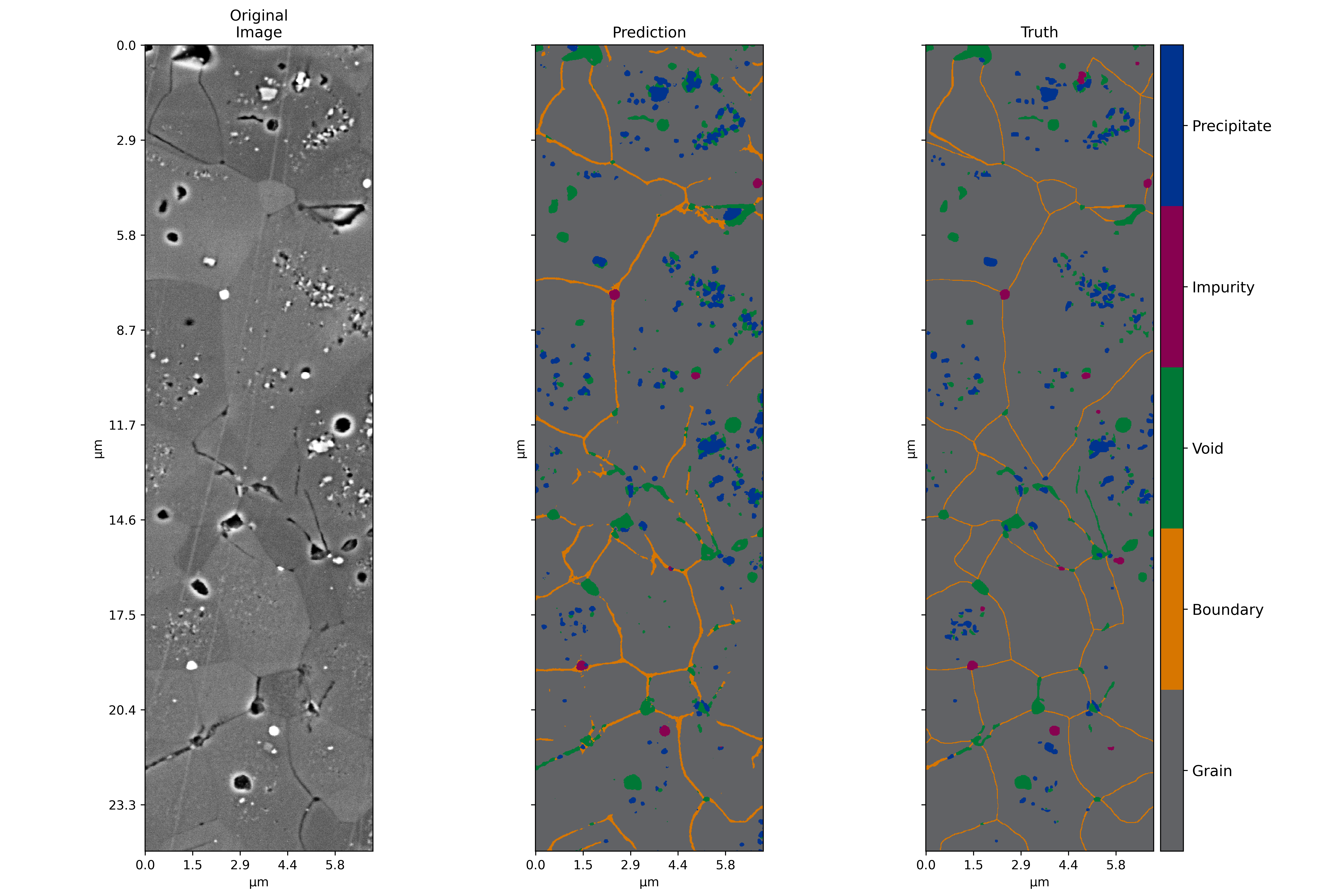}
      A\caption{\label{fig:S4_unirradiated_image_4} Predictions: From left to right, 1) the original test section of ``Image 4'' which was irradiated, 2) the predicted image, and 3) the expert-labeled image. The predicted images were created with these parameters: Small Bayesian SegNet model architecture, EWCE loss, and Adam optimizer.}
    
\end{figure}

\begin{figure}[htbp]
    
    \includegraphics[width=\textwidth]{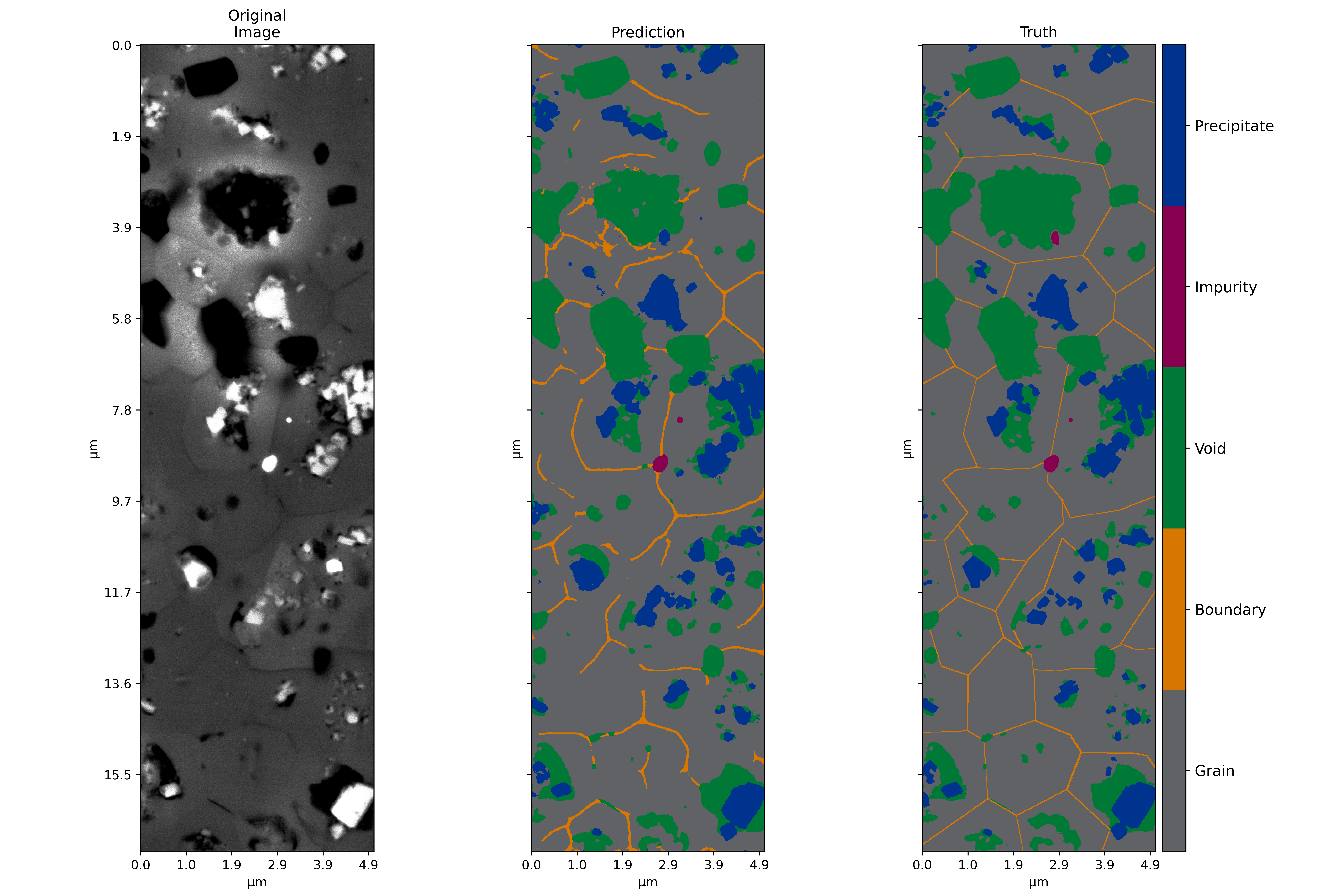}
      \caption{\label{fig:S5_unirradiated_image_5} Predictions: From left to right, 1) the original test section of ``Image 5'' which was irradiated, 2) the predicted image, and 3) the expert-labeled image. The predicted images were created with these parameters: Small Bayesian SegNet model architecture, EWCE loss, and Adam optimizer.}
    
\end{figure}
 \clearpage

\end{document}